\documentclass[]{interact}

\usepackage{epstopdf}
\usepackage[caption=false]{subfig}

\usepackage[numbers,sort&compress]{natbib}
\bibpunct[, ]{[}{]}{,}{n}{,}{,}

\usepackage{mathrsfs, mathtools, nicefrac}
\usepackage{graphicx, tikz}
\usepackage{algpseudocode, algorithm}
\usepackage{microtype}
\usepackage{enumitem}
\usepackage{hyperref, cleveref}
\usepackage{mleftright} 
    \mleftright

\newcommand{\cF}{\mathcal{F}}

\newcommand{\cM}{\mathcal{M_{\lno \cdot}}}
\newcommand{\cN}{\mathcal{N}_{\lno \cdot}}

\newcommand{\cX}{\mathcal{X}}

\newcommand{\I}{\mathbb{I}}

\newcommand{\R}{\mathbb{R}}

\newcommand{\e}{\varepsilon}

\newcommand{\lrb}[1]{\left(#1\right)}
\newcommand{\brb}[1]{\bigl(#1\bigr)}
\newcommand{\Brb}[1]{\Bigl(#1\Bigr)}

\newcommand{\lcb}[1]{\left\{#1\right\}}
\newcommand{\bcb}[1]{\bigl\{#1\bigr\}}
\newcommand{\Bcb}[1]{\Bigl\{#1\Bigr\}}

\newcommand{\lce}[1]{\left\lceil#1\right\rceil}
\newcommand{\bce}[1]{\bigl\lceil#1\bigr\rceil}

\newcommand{\labs}[1]{\left\lvert#1\right\rvert}
\newcommand{\babs}[1]{\bigl\lvert#1\bigr\rvert}
\newcommand{\Babs}[1]{\Bigl\lvert#1\Bigr\rvert}

\newcommand{\lno}[1]{\left\lVert#1\right\rVert}
\newcommand{\bno}[1]{\bigl\lVert#1\bigr\rVert}

\DeclareMathOperator*{\argmax}{argmax}
\newcommand{\dif}{\,\mathrm{d}}
\newcommand{\lip}{Lipschitz}
\newcommand{\lipn}{Lipschitzness}
\newcommand{\leb}{Lebesgue}
\newcommand{\bx}{\boldsymbol{x}}
\newcommand{\bxs}{\boldsymbol{x}^{\star}}
\newcommand{\by}{\boldsymbol{y}}
\newcommand{\bzero}{\boldsymbol{0}}
\newcommand{\s}{\subset}

\newcommand{\fracc}[2]{#1/#2}

\newcommand{\piya}{Piyavskii--Shubert algorithm}
\newcommand{\CpiyA}{Certified Piyavskii--Shubert algorithm}
\newcommand{\AcpiyA}{Approximate Certified Piyavskii--Shubert Algorithm}
\newcommand{\apiya}{Approximate Piyavskii--Shubert algorithm}

\renewcommand{\epsilon}{\varepsilon}
\newcommand{\eo}{\varepsilon_0}
\newcommand{\me}{m_{\varepsilon}}

\newcommand{\fh}{\widehat{f}}
\newcommand{\fhs}{\widehat{f}^{\star}}
\newcommand{\ds}{d^{\star}}
\newcommand{\Cs}{C^{\star}}
\newcommand{\xs}{x^{\star}}
\newcommand{\Xe}{\cX_\e}

\newcommand{\SNC}{S_{\mathrm{NC}}}
\newcommand{\SC}{S_{\mathrm{C}}}
\newcommand{\vol}{\mathrm{vol}}
\newcommand{\diam}{\mathrm{diam}_{\lno \cdot}}
\newcommand{\ball}{B_{\lno \cdot}}
\newcommand{\mink}{Minkowski}
\newcommand{\Lip}{\mathrm{Lip}}

\newcommand{\piy}{Piyavskii--Shubert}

\theoremstyle{plain}
\newtheorem{theorem}{Theorem}[section]
\newtheorem{lemma}[theorem]{Lemma}

\newtheorem{proposition}[theorem]{Proposition}

\theoremstyle{definition}

\newtheorem{assumption}[theorem]{Assumption}

\theoremstyle{remark}

\begin{document}


\title{Regret analysis of the Piyavskii-Shubert algorithm for global Lipschitz optimization}

\author{
\name{%
    Cl\'{e}ment Bouttier\textsuperscript{a}
    Tommaso Cesari\textsuperscript{b,c}\thanks{CONTACT Tommaso Cesari. Email: tommaso.cesari@tse-fr.eu},
    M\'elanie Ducoffe\textsuperscript{d}, and
    S\'{e}bastien Gerchinovitz\textsuperscript{d,e}%
}
\affil{%
    \textsuperscript{a}AURA AERO, Toulouse, France;
    \textsuperscript{b}Universit\`a degli Studi di Milano, Milano, Italy;
    \textsuperscript{c}Toulouse School of Economics, Toulouse, France;
    \textsuperscript{d}IRT Saint Exup\'ery, Toulouse, France; 
    \textsuperscript{e}Institut de Math\'ematiques de Toulouse, Toulouse, France.%
}
}

\maketitle

\begin{abstract}
We consider the problem of maximizing a non-concave Lipschitz multivariate function over a compact domain by sequentially querying its (possibly perturbed) values. We study a natural algorithm designed originally by Piyavskii and Shubert in 1972, for which we prove new bounds on the number of evaluations of the function needed to reach or certify a given optimization accuracy. Our analysis uses a bandit-optimization viewpoint and solves an open problem from Hansen et al.\ (1991) by bounding the number of evaluations to certify a given accuracy with a near-optimal sum of packing numbers. 
\end{abstract}



\section{Introduction}

In this paper, $f\colon \cX \to \R$ denotes an unknown function defined on a known compact subset $\cX$ of $\R^d$ containing at least two points (if $\labs{\cX}\le 1$, the problem becomes trivial). 
At a high level, we consider the following global optimization problem: with only black-box access to $f$, find an  approximate maximizer of $f$ at prescribed accuracy $\epsilon$, with as little evaluations of $f$ as possible. We make the setting more precise below.

\subsection{Setting: Lipschitz black-box optimization, with or without certificates}
\label{s:settingintro}

We make the following weak Lipschitz assumption,\footnote{\Cref{ass:base} is sometimes referred to as \emph{local smoothness} \citep{munos2014bandits} and close in spirit to \emph{calmness} \citep{ROCKAFELLAR1985665}.
} where the norm $\lno{\cdot}$ and the constant $L$ are known to the learner. For some of our results, we will instead assume $f$ to be globally $L$-Lipschitz.

\begin{assumption}[Lipschitzness around a maximizer]
\label{ass:base}
The function $f$ attains its maximum at some $\bxs \in \cX$, and there exist a constant $L > 0$ and a norm $\lno{\cdot}$ such that, for all $\bx \in \cX$,
\[
	f(\bx)
\ge 
	f(\bxs) - L \lno{ \bxs - \bx }\;.
\]
\end{assumption}

\paragraph*{Online learning protocol.}

We study the case in which $f$ is black-box, i.e., except for some \emph{a priori} knowledge on its smoothness, we can only access $f$ by sequentially querying its values at a family $\bx_1,\bx_2,\ldots \in \cX$ of points of our choice.
At every round $k \ge 1$, the query point $\bx_k$ can be chosen as a deterministic function of the values $f(\bx_1),\ldots,f(\bx_{k-1})$ observed so far. 
At the end of round $k$, using the values $f(\bx_1),\ldots,f(\bx_{k})$, the learner outputs a recommendation $\bxs_k \in \cX$ with the goal of minimizing the \emph{optimization error} (or \emph{simple regret}): $\max_{\bx \in \cX} f(\bx) - f(\bxs_k) = f(\bxs) - f(\bxs_k)$.

The learner may also output an \emph{error certificate}, depending on whether the algorithm is certified or not. More precisely, in all the sequel, we consider two different types of algorithms:
\begin{itemize}
    \item \emph{Non-certified algorithms} only output a recommendation $\bxs_k \in \cX$ at the end of each time step $k$.
	\item \emph{Certified algorithms} output an \emph{error certificate} $\xi_k \ge 0$ together with the recommendation $\bxs_k$ at the end of each time step $k$. 
	The error certificate has to satisfy
	\[
	    f(\bxs)-f(\bxs_k) 
    \le
        \xi_k
    \]
    for any function $f$ that is $L$-\lip{} around $\bxs$. 
    In other words, the error certificate guarantees that the recommendation $\bxs_k$ is $\xi_k$-optimal.
\end{itemize}

These two types of algorithms are summarized in Online~Protocol~\ref{alg:base} below.

{
\makeatletter
\renewcommand*{\ALG@name}{Online Protocol}
\begin{algorithm}
\caption{\label{alg:base} Non-certified (/\emph{certified}) setting}
\begin{algorithmic}
\For
{%
    $k = 1, 2, \ldots$
}
    \State pick the next query point $\bx_k \in \cX $\;
    \State observe the value $f(\bx_k)\in \R$\;
    \State output a recommendation $\bxs_k \in \cX$\;
    \State output an error certificate $\xi_k \ge 0$ (\emph{certified setting only})\;
\EndFor
\end{algorithmic}
\end{algorithm}
\makeatother
}

\paragraph*{The \piya{}.} Several algorithms were designed for these settings in the past (see related works in Section~\ref{s:relatedworks}). 
In this paper, we focus on arguably one of the most natural ones, first appearing in the one-dimensional works of Piyavskii \cite{Piy-72-AbsoluteExtremum} and Shubert \cite{shubert1972sequential}. 
The principle is simple: at each round $k$, the query point $\bx_k$ is chosen as a (possibly approximate) maximizer of the piecewise-conic proxy function
\begin{align*}
    \fh_{k-1} \colon \cX  & \to \R \\
    \bx               & \mapsto \min_{i\in\{1,\dots,k-1\}} \bcb{ f(\bx_i) + L \lno{\bx_i - \bx}}
\end{align*}
(see Figure~\ref{fig:PYbehavior}), which satisfies $\fh_{k-1}(\bxs) \ge f(\bxs)$ under Assumption~\ref{ass:base}, and is the tightest upper bound on $f$ given the available information when $f$ is $L$-globally Lipschitz.
At the end of round $k$, the error certificate is chosen as $\xi_k = \max_{\bx \in \cX} \widehat f_k(\bx) - f(\bxs_k)$, which upper bounds $f(\bxs) - f(\bxs_k)$ by construction.

In this paper, we study the sample complexity of both the non-certified and certified versions of the \piya{}. That is, for any target accuracy $\epsilon>0$, we bound the number $n$ of evaluations of $f$ after which $f(\bxs) - f(\bxs_n) \le \epsilon$ (i.e., the recommendation of the algorithm is $\epsilon$-optimal) or after which $\xi_n \le \epsilon$ (i.e., the algorithm can certify from the observed data that the recommendation is $\epsilon$-optimal). As will become apparent from our bounds, certifying the accuracy of a recommendation is sometimes much harder than just outputting an approximate maximizer with correct (but unknown) accuracy (see comment after \Cref{t:SC-upper}).

\paragraph*{Open questions.} Though the \piya{} is very natural, little is known about its sample complexity. In the non-certified setting, we are only aware of the worst-case bound $f(\bxs) - f(\bxs_n) \lesssim n^{-1/d}$, which corresponds to the fact that $f(\bxs) - f(\bxs_n) \le \epsilon$ whenever $n \gtrsim (1/\epsilon)^d$. In the certified setting, Hansen et al.\ \cite{hansen1991number} proved in dimension $d=1$ that $\xi_n \leq \epsilon$ whenever $n \gtrsim \int_0^1 \brb{ f(\xs)-f(x)+\epsilon }^{-1} \dif x$, but the authors claimed that ``\emph{extending the results of this paper to the multivariate [case] appears to be difficult}''. We refine or extend these previous bounds in the sequel. See below and the related works in Section~\ref{s:relatedworks} for more details.

\subsection{Outline of the paper and main contributions}
\label{s:contributions}

In \Cref{s:ass-def-not}, we present some recurring notation.
We recall the definition of the \piya{} in \Cref{s:defPS}. 
We then make the following contributions:
\begin{itemize}
    \item In Sections~\ref{s:lemmas} and~\ref{s:upperbounds} we derive optimization error bounds for the \piya{} in any dimension $d \geq 1$, both in the non-certified and the certified settings (\Cref{t:SNC-upper,t:SC-upper}). 
    We bound the sample complexity in terms of packing numbers of sets of suboptimal points at different scales. 
    As mentioned above, error bounds for the \piya{} were previously known only for worst or special cases. 
    Moreover, the absence of any underlying discrete structure (as opposed to, e.g., the DOO algorithm \citep{Per-90-OptimizationComplexity,munos2011optimistic}, or, to some extent, the Zooming algorithm \cite{kleinberg2008multi}) makes the \piya{} and its analysis quite simple. We hope this helps better understand the role of packing numbers in such types of results.
    \item In Sections~\ref{s:optimality} and~\ref{s:integralrepresentation} we reinterpret the error bounds of Section~\ref{s:upperbounds} in view of the very recent results of Bachoc et al.\ \cite{bachoc21-ZerothOrderLipschitzOptimization}.%
    \footnote{For contingent reasons, \cite{bachoc21-ZerothOrderLipschitzOptimization} was written after but submitted before the present paper.
    Since there are some dependencies between the results of these two papers, we clarify in \Cref{s:relatedworks} which result belongs to which paper, and that no logic loops arise from cross-citing.} 
    The latter results have two consequences: (i) our error bound in the certified case is nearly instance-optimal; (ii) this bound is proportional to the integral $\int_\cX \mathrm{d}\bx/\brb{ f(\bxs) - f(\bx) + \e }^d$ for any dimension $d \geq 1$, which solves the question left open by Hansen et al.\ \cite{hansen1991number} three decades ago. 
    \item In Section~\ref{s:robustness} we investigate the robustness of the \piya{} to deterministic perturbations and imperfect maximization of the proxy function. We prove optimization error bounds that generalize those of Section~\ref{s:upperbounds}, by allowing both perturbed evaluations of $f$ and imperfect maximization of the proxy function.
\end{itemize}
Related works from the global optimization and bandit optimization literatures are discussed in details in Section~\ref{s:relatedworks}. 
Finally, although the main proofs appear in the main body, some technical details are postponed to the appendix, where we also recall some useful results on packing and covering numbers.

\subsection{Recurring notation}
\label{s:ass-def-not}

We begin by recalling the definition of packing numbers, which is of utmost importance for our analysis. 
In \Cref{sec:lemmaPacking}, we also recall a few known inequalities about packing (and covering) numbers that will prove useful throughout the paper.

Let $\lno \cdot$ be a norm.
For any bounded set $A\s\R^d$ and $r>0$, the $r$-\emph{packing number} of $A$ (with respect to $\lno \cdot$) is the largest number of $r$-separated points contained in $A$, i.e.,
\[
	\cN(A,r)
:= 
	\sup \lcb{
		k \in \{1,2,\dots\}  :  \exists \bx_1, \dots, \bx_k \in A, \min_{i\neq j} \lno{ \bx_i - \bx_j } > r
	} \;,
\]
if $A$ is nonempty, zero otherwise.

Our results will be expressed in terms of packing numbers of near-optimal points and layers, defined as follows.
Let $\cX \s \R^d$, $f\colon \cX \to \R$, and $\bxs$ be a maximizer of $f$.
For all $\e > 0$, we denote the set $\bcb{ \bx \in  \cX : f(\bx) \ge f(\bxs)-\e }$ of all $\e$-\emph{optimal points} of $f$ by $\cX_\e$ and its complement (i.e., the set of $\e$-\emph{suboptimal points}) by $\cX_\e^c$.
For all $0\le a<b$, we denote the $(a,b]$-\emph{layer} $\cX_a^c \cap \cX_b$ (i.e., the set of points that are $a$-suboptimal but $b$-optimal) by $\cX_{(a,\, b]}$.

Note that if $f\colon \cX \to \R$ is $L$-\lip{} around a maximizer with respect to a norm $\lno \cdot$, then every point in $\cX$ is $\eo$-optimal, where 
$
	\eo 
=
    \eo \brb{ \cX,L,\lno \cdot }
:=
	L \sup_{\bx,\by \in  \cX } \lno{ \bx-\by }$.
In other words, $\cX_{\eo}= \cX$. 
For this reason, without loss of generality we will only consider values of $\e$ smaller than $\eo$.

For all $x \in \R$, we denote the smallest integer greater than or equal to $x$ by $\lce x$. We also write $[k] = \{1,\ldots,k\}$ for any integer $k \ge 1$.
For any norm $\lno \cdot$, radius $r>0$, and center $\bx \in \R^d$, we denote by $\ball(r,\bx)$ the ball in $\brb{ \R^d, \lno \cdot }$ of radius $r$ centered at $\bx$.
For any norm $\lno \cdot$ and set $E$, we denote the diameter $\sup_{\bx,\by\in\cX}\lno{\bx-\by}$ of $E$ (with respect to $\lno \cdot$) by $\diam(E)$.
We denote the \leb{} measure of a (\leb{}-measurable) set $E \s \R^d$ by $\vol(E)$ and we simply refer to it as its \emph{volume}.

\section{The \piya{}}
\label{s:defPS}

\begin{figure}
\centering
\begin{tikzpicture}[scale=0.5]

\draw (4,3.5) node[left, blue] {$f$};

\draw plot coordinates {(4,6) (5,5) (7.45,7.45)};
\draw[dotted] plot coordinates { (7.45,7.45)(12,12)};
\draw[dotted] plot coordinates {(8.5,8.5) (9.5,7.43)};
\draw[dotted] plot coordinates{(0,10) (4,6) };
\draw plot coordinates {(9.5,7.43)(12,5)};
\draw plot coordinates {(0,2) (4,6)};
\draw plot coordinates {(7.45,7.45) (8.5,6.43)(9.5,7.43)};

\filldraw [gray] (12,5) circle (2pt);

\filldraw [gray] (5,5) circle (2pt);
\draw[dotted] (5,0) -- (5,5);
\draw (5,0) node[below] {$x_1$};
\draw (5,4.5) node[right] {$y_1$};

\filldraw [gray] (8.5,6.43) circle (2pt);
\draw[dotted] (8.5,0) -- (8.5,8.5);
\draw (8.5,0) node[below] {$x_4$};
\draw (8.2,8.5) node[left] {$\max(\hat{f}_3)$};
\draw (8.5,6) node[right] {$y_4$};
\draw (8.5,8.5) circle (3pt);

\draw (0,0) node[below right] {$x_3$};
\draw (0.1,10) node[right] {$\max(\hat{f}_2)$};
\draw (0,2) node[left] {$y_3$};
\draw (0,10) circle (3pt);

\draw (12,0) node[below left] {$x_2$};
\draw (11.8,12) node[left] {$\max(\hat{f}_1)$};
\draw (12,5) node[left] {$y_2$};
\draw (12,12) circle (3pt);

\draw[dotted] (7.45,0) -- (7.45,7.45);
\draw (7.45,0) node[below] {$x_5$};
\draw (7.2,7.45) node[left] {$\max(\hat{f}_4)$};

\draw (7.45,7.45) circle (3pt);

\draw (4,6) circle (3pt);
\draw (9.5,7.43) circle (3pt);

\filldraw [gray] (0,2) circle (2pt);
\draw[thick, blue] (0,2) .. controls (1,1.6) and (1,2) .. (2,3);
\draw[thick, blue] (2,3) .. controls (3,4) and (4,5) .. (5,5);
\draw[thick, blue] (5,5) .. controls (7,5) and (7,5) .. (8,6);
\draw[thick, blue] (8,6) .. controls (9,7) and (10,7) .. (12,5);
\draw[->] (-2,0) -- (14,0);

\draw (12, 0) node[below right] {$1$};
\draw (0,0) node[below left] {$0$};
\draw [dotted] (0,-1) -- (0,13);

\draw [dotted] (12,-1) -- (12,13);
\end{tikzpicture}
\caption{\label{fig:PYbehavior} First four iterations of the \piya{} in dimension $d=1$. In thick blue the function $f$. In black the proxy function $\fh_4$.}
\end{figure}

The \piya{} (\Cref{a:cpiya}, illustrated in \Cref{fig:PYbehavior}) works by maintaining a proxy $\fh_k$ of the objective function $f$ that upper bounds $f$ at the
maximizer $\bxs$ (see \Cref{l:fh}). 
The next query point $\bx_{k+1}$ is chosen as a maximizer of $\fh_k$. 
The intuition is that $\bx_{k+1}$ is either an approximate maximizer of $f$ (which reduces the optimization error immediately) or belongs to a largely unexplored suboptimal region (which helps reduce the regret in the future, see \Cref{l:good-exploration,l:spread-exploration}). 
Note the similarity with the ``optimism in
the face of uncertainty'' principle in stochastic multiarmed bandits \cite{LS18-banditalgos}. 
There,
a proxy for the real mean of an arm is given by an upper confidence bound, and arms can have high upper
confidence bounds only if they are either actually good, or they belong to an under-explored set of arms.
In \Cref{s:lemmas}, we will show that the certified version of the \piya{} is indeed a certified algorithm.
\begin{algorithm}
\caption{\piy{}, for the non-certified (/\emph{certified}) setting}
\label{a:cpiya}
\textbf{input:} \lip{} constant $L>0$, norm $\lno{\cdot}$, initial guess $\bx_1 \in \cX$
\begin{algorithmic}
\For{$k = 1, 2, \dots$}
	\State pick the next query point $\bx_k$
    \State observe the value $f(\bx_k)$
    \State output the recommendation $\bxs_k \gets \argmax_{\bx\in\{\bx_1, \ldots, \bx_k\}} f(\bx) $
    \State update the proxy function
    \[
        \widehat f_k(\cdot) \gets \min_{i\in[k]} \Bcb{ f(\bx_i) + L \bno{ \bx_i - (\cdot) } }
    \]
    \State output the error certificate $\xi_k \gets \widehat f^\star_{k} - f^\star_{k}$, where
    \[
        \widehat f ^\star_k \gets \max_{\bx \in \cX} \widehat f_k(\bx) \;, \quad
        f^\star_k \gets \max_{i\in[k]} f(\bx_i)
        \quad \text{(\emph{certified setting only})}
    \]
    \State let $\bx_{k+1} \in \argmax_{\bx \in \cX} \widehat f_k(\bx)$
\EndFor
\end{algorithmic}
\end{algorithm}

It is worth noting that the \piya{} might not be computationally efficient in high dimensions. 
Indeed, for a budget of $n$ evaluations of 
$f$, the problem of optimizing $f$ is replaced by the optimization of the $n$ proxy functions $\fh_1,\ldots,\fh_n$, which could be a demanding task on its own.
To see this, note that finding the maximum of $\fh_k$ is related to the problem of determining a Voronoi diagram for a set of $k$ points, which is known to have a computational complexity that is exponential in the dimension of the ambient space \cite[Theorem 4.5 , Section 4.3.2 ``Power diagrams and convex hulls'']{aurenhammer2000voronoi}. 
Some algorithms get around these computational issues by replacing the piecewise-conic proxies of the \piya{} with looser but simpler functions. 
The DOO algorithm \cite{munos2011optimistic} (see also \cite{Per-90-OptimizationComplexity}), for example, uses piecewise-constant functions, with pieces that correspond to a predetermined hierarchical partition of $\cX$. 

Beyond pedagogical reasons and the fact that it appears to be one of the most natural methods in our setting, the \piya{} can be applied to low-dimensional real-life problems (e.g., hyperparameter tuning), but also to maximization problems in which the computational cost is driven by the evaluation of the objective~$f$ rather than the dimension of the ambient space. Indeed, when measuring algorithms by their sample complexity (number of evaluations of $f$) rather than their running time (number of elementary operations aside evaluations of $f$), we conjecture that the \piya{} is better (by non-negligible multiplicative constants) than simplified algorithms such as DOO. This is supported by some low-dimensional experiments carried out in \cite{hansen1992global2}.

\section{Error bounds}

We begin this section by showing some key properties of the \piya{} (\Cref{s:lemmas}).
We then prove upper bounds on the number of queries that the \piya{} needs before being able to reach or certify a certain accuracy (\Cref{s:upperbounds}).
Afterwards, we discuss optimality (\Cref{s:optimality}), and give a compact integral representation of our upper bound in the certified setting (\Cref{s:integralrepresentation}).

\subsection{Useful lemmas}
\label{s:lemmas}

We begin this section by proving two important properties of the proxy functions $\fh_k$ that hold whenever the objective function $f$ is \lip{} around a maximizer $\bxs$.
First, the proxy functions are always greater than or equal to the objective function at the maximizer $\bxs$.
(Note that when $f$ is globally \lip{}, the lower bound $\fh_k(\bx) \ge f(\bx)$ is well known and true not only for $\bx=\bxs$ but for all $\bx \in \cX$.) 
Second, the proxy functions are always lower than or equal to the objective function at all past query points (so, in particular, the same is true for all recommendations).

\begin{lemma}
\label{l:fh}
Assume that $f\colon \cX \to \R$ is $L$-\lip{} around a maximizer $\bxs$ with respect to a norm $\lno \cdot$ (\Cref{ass:base})
and the \piya{} (\Cref{a:cpiya}) is run with input $L$, $\lno \cdot$, $\bx_1$.
Then, for all $k \ge 1$, the proxy function 
$
    \fh_{k}(\cdot)
=
    \min_{i \in [k]} \bcb{ f(\bx_i)+L \lno{ \bx_i-(\cdot) } }
$
is $L$-Lipschitz and satisfies
\[
	\fh_k(\bxs) \ge f(\bxs) 
\qquad \text{ and } \qquad 
	\fh_k(\bx_j) \le f(\bx_j) \;, \quad \forall j\in[k] \;.
\]
\end{lemma}	

\begin{proof}
Fix any $k\ge 1$. 
The fact that $\fh_k$ is $L$-\lip{} is straightforward. 
Moreover, by the $L$-\lipn{} of $f$ around $\bxs$, we get
\begin{align*}
    \fh_{k}(\bxs)
&
=
    \min_{i \in[k]} \Bcb{ f(\bx_i) + L \lno{ \bx_i-\bxs } }
\\
&
\ge
    \min_{i \in[k]} \Bcb{ \brb{ f(\bxs) - L \lno{ \bx_i-\bxs } } + L \lno{ \bx_i-\bxs } }
=
    f(\bxs) \;.
\end{align*}
Furthermore, for all $j \in [k]$, the definition of $\fh_k(\bx_j)$ implies immediately
\[
	\fh_k(\bx_j) 
= 
	\min_{i \in[k]} \Bcb{ f(\bx_i) + L \lno{ \bx_i-\bx_j } }
\le
	f(\bx_j) + L \lno{ x_j - x_j }
=
	f(\bx_j) \;. \qedhere
\]
\end{proof}
A first immediate consequence of \Cref{l:fh} is that the certified version of the \piya{} is indeed a certified algorithm, as defined in \Cref{s:settingintro}.

\begin{lemma}
Assume that $f\colon \cX \to \R$ is $L$-\lip{} around a maximizer $\bxs$ with respect to a norm $\lno \cdot$ (\Cref{ass:base})
and the \CpiyA{} (\Cref{a:cpiya}) is run with input $L$, $\lno \cdot$, $\bx_1$.
Then, for all $n\ge 1$, $f(\bxs)-f(\bxs_n) \le \xi_n$.
\end{lemma}

\begin{proof}
By \Cref{l:fh} and the definitions of $f^\star_n$, $\fhs_n$, and $\xi_n$, for all $n\ge 1$, we get
$
    f(\bxs) - f(\bxs_n) 
\le
    \fh_n(\bxs) - f^\star_n
\le
    \xi_n
$.
\end{proof}

Another important implication of \Cref{l:fh} is that if the \piya{} observes $f$ at a $\Delta$-suboptimal point $\bx_i$, then all the next query points $\bx_j$ are at least $(\Delta/L)$-distant from $\bx_i$. 
In other words, the algorithm does not explore too much in suboptimal regions.

\begin{lemma}
\label{l:good-exploration}
Assume that $f\colon \cX \to \R$ is $L$-\lip{} around a maximizer $\bxs$ with respect to a norm $\lno \cdot$ (\Cref{ass:base})
and the \piya{} (\Cref{a:cpiya}) is run with input $L$, $\lno \cdot$, $\bx_1$.
Fix any $\Delta>0$ and assume that there exists $i \ge 1$ such that the $i$-th query point $\bx_i$ belongs to $\cX_{\Delta}^c$. 
Then, for all $j > i$, the $j$-th query point satisfies
\[
	\lno{ \bx_j-\bx_i }
>
    \frac \Delta L \;.
\]
\end{lemma} 

\begin{proof}
Fix any $j>i$. 
Then, using the fact that $\bx_j$ has been selected as the $j$-th query point (in the first inequality), \Cref{l:fh} (in the second and fourth), and $\bx_i \in \cX_\Delta^c$ (in the third), we get
\[
	\fh_{j-1}(\bx_j)
\ge 
	\fh_{j-1}(\bxs)
\ge 
	f(\bxs)
>
	f(\bx_i)+\Delta
\ge 
	\fh_{j-1}(\bx_i)+\Delta \;.
\]
Since $\fh_{j-1}$ is $L$-\lip{} (by \Cref{l:fh}), we can thus conclude that
$
    L \lno{ \bx_j-\bx_i} 
\ge 
    \babs{ \fh_{j-1}(\bx_j)-\fh_{j-1}(\bx_i) }
>
    \Delta
$.
\end{proof}

As a final corollary to \Cref{l:fh}, we show that the \CpiyA{} automatically adapts its exploration to all possible approximation scales at the same time. 
In words, for any accuracy $\e$, until the algorithm can certify that the optimization error is $\e$ or better (i.e., until the error certificate is smaller than or equal to $\e$), it only queries points that are at least $(\e/L)$-distant from all past query points.
The strength of this statement comes from the fact that the algorithm \emph{does not} take an accuracy $\e$ as a parameter and the result is consequently not proved for a specific and known $\e$: it holds for all $\e$ simultaneously.

\begin{lemma}
\label{l:spread-exploration}
Assume that $f\colon \cX \to \R$ is $L$-\lip{} around a maximizer $\bxs$ with respect to a norm $\lno \cdot$ (\Cref{ass:base})
and the \CpiyA{} (\Cref{a:cpiya}) is run with input $L$, $\lno \cdot$, $\bx_1$. 
Fix any $\e > 0$ and let $\tau$ be the first time where the algorithm returns an error certificate $\xi_\tau \le \e$.
Then, for all distinct $i,j \in [\tau]$, 
\[
    \lno{ \bx_i - \bx_j } 
> 
    \frac{\e}{L}
    \;.
\]
\end{lemma}

\begin{proof}
Without loss of generality, assume $1 \le i < j \le \tau$. 
Then, by the definition of $\fh_{j-1}^\star$ (first equality), that of $\xi_{j-1}$ together with the assumption $\xi_{j-1} > \e$ (first inequality), the definition of $f_{j-1}^\star$ (second), and \Cref{l:fh} (last), we get
\[
    \fh_{j-1}(\bx_j)
=
    \brb{  \fh_{j-1}^\star - f_{j-1}^\star }
    +
    f_{j-1}^\star
>
    \e + f_{j-1}^\star
\ge
    \e + f(x_i)
\ge
    \e + \fh_{j-1}(\bx_i) \;.
\]
Since $\fh_{j-1}$ is $L$-\lip{} (by \Cref{l:fh}), we can thus conclude that
$
    L \lno{ \bx_j-\bx_i} 
\ge 
    \babs{ \fh_{j-1}(\bx_j)-\fh_{j-1}(\bx_i) }
>
    \e
$.
\end{proof}

The previous result shows that $\xi_n$ serves a synergistic double-role in the \CpiyA{}.
When it is large, it guarantees that the queries are sufficiently sparse, and as soon as it becomes small, it certificates that the recommendation $\bx_n$ is sufficiently good.

\subsection{Upper bounds}
\label{s:upperbounds}

We now prove an upper bound on the number of evaluations of $f$ that are needed before the recommendations of the \piya{} become near-optimal.
To this end, we use a peeling technique in which the input space $\cX$ is partitioned in terms of the output values of $f$ and leverage the fact that highly suboptimal regions $\cX_{(\e_k, \e_{k-1}]}$ (with $\e_k \gg \e$) need not be explored too much thanks to the results we showed in \Cref{s:lemmas}.

\begin{theorem}
\label{t:SNC-upper}
Assume that $f\colon \cX \to \R$ is $L$-\lip{} around a maximizer $\bxs$ with respect to a norm $\lno \cdot$ (\Cref{ass:base}),
$\eo := L \sup_{\bx,\by \in  \cX } \lno{ \bx - \by }$, $\e \in (0,\eo)$, $\me := \bce{ \log_2(\eo/\e) }$, $\e_{\me} := \e$, $\e_k := \eo 2^{-k}$ for all $k < \me$, and
\[
	\SNC(f,\e)
:= 
	\sum_{k=1}^{\me} 
	\cN\lrb{ \cX_{(\e_k ,\, \e_{k-1}]} ,\, \frac{\e_k}{L} }\;.
\]
If $n \ge 1+\SNC(f,\e)$, then the $n$-th recommendation $\bxs_n$ of the \piya{} (\Cref{a:cpiya}, run with input $L$, $\lno \cdot$, $\bx_1$) satisfies $f(\bxs)-f(\bxs_n) \le \epsilon$.
\end{theorem}

We can draw immediate corollaries from \Cref{t:SNC-upper} if $f$ is such that $\cN(\cX_r ,\, r) \le \Cs / r^{\ds}$, for some $\Cs>0$, $\ds\in[0,d]$, and all $r \in (\e,\eo]$ (i.e., if the \emph{near-optimality dimension} is $\ds$ ---see the related works in \Cref{s:relatedworks} for more details).
In this case, $\SNC(f,\e) \lesssim \log(1/\e)$ if $\ds=0$, and $\SNC(f,\e) \lesssim 1/\e^{\ds}$ if $\ds>0$.
Examples of these functions are linear functions of the norms $f(\bx) = a - b \lno{\bx - \bx_0 }$, where $\ds=0$, or quadratic functions of the norms $f(\bx) = a - b \lno{\bx - \bx_0 }^2$, where $\ds=d/2$.

Importantly, our bound in terms of $\SNC(f,\e)$ is more general than the previous corollaries. It  accommodates cases in which $f$ has different of such $\ds$'s at different scales $r$,
e.g., the function 
$
    f(\bx) 
= 
    \brb{ 1/4 - \lno{\bx}^2 } \I \bcb{ \lno {\bx} \le 1/2 } 
    + 
    \brb{ 1/2 - \lno{\bx} } \I \bcb{ \lno {\bx} > 1/2 } 
$,
for which $\SNC(f,\e) \lesssim \log(1/\e)$ for large values of $\e$ (linear regime if $\e \ge 1/2$) but $\SNC(f,\e) \lesssim 1/\e^{d/2}$ for small values of $\e$ (quadratic regime if $\e < 1/2$). 
Many different examples can be designed this way.

\begin{proof}
By definition of $\e_k$, we can partition the set $\cX_\e^c = \cX_{(\e ,\, \eo]}$ of $\e$-suboptimal points into $\me$ layers
\[
    \cX_\e^c = \bigcup_{k = 1}^{\me} \cX_{(\e_k ,\, \e_{k-1}]} \;.
\]
Therefore, for any $i \ge 1$, if $\bx_i\in\Xe^c$, then there exists a unique $k\in [\me]$ such that $\bx_i \in \cX_{(\e_k ,\, \e_{k-1}]}$.
Since the \piya{} does not explore too much in suboptimal regions, \Cref{l:good-exploration} and the definition of packing numbers imply that for all $k\in [\me]$, the maximum number of times that a query point $\bx_i$ is chosen in $\cX_{(\e_k ,\, \e_{k-1}]} \s \cX_{\e_k}^c$ is upper bounded by 
$
    \cN \lrb{ \cX_{(\e_k ,\, \e_{k-1}]} ,\, \frac{\e_k}{L} }
$.
Hence
\begin{align*}
    \Babs{ \bcb{ i\in\{1,2,\dots\} : \bx_i\in\Xe^c } }
&
=
    \sum_{k=1}^{\me} \Babs{ \bcb{ i\in\{1,2,\dots\} : \bx_i\in \cX_{(\e_k ,\, \e_{k-1}]} } }
\\
&
\le
    \sum_{k=1}^{\me}\cN \lrb{ \cX_{(\e_k ,\, \e_{k-1}]} ,\, \frac{\e_k}{L} }
=
    \SNC(f,\e).    
\end{align*}
Therefore, if
$
    n
\ge
    1 + \SNC(f,\e)
$,
then there exists $i\in[n]$ such that $\bx_i\in\Xe$, which implies (by definition of $\bxs_n$) that 
$
    f(\bxs_n)
\ge 
    f(\bx_i)
\ge 
    f(\bxs) - \epsilon
$ 
and concludes the proof.
\end{proof}

We now prove the analogous (in the certified setting) of the previous result.
Its proof combines the ideas in \Cref{t:SNC-upper} with the synergistic double-role of the error certificate presented in \Cref{l:spread-exploration}.

Notably, when $n\ge \SC(f,\e)$, not only $f(\bxs)-f(\bxs_n) \le \e$ but the \CpiyA{} can certify that this bound holds at run-time based only on the available data. See below the proof for further comments.

\begin{theorem}
\label{t:SC-upper}
Assume that $f\colon \cX \to \R$ is $L$-\lip{} around a maximizer $\bxs$ with respect to a norm $\lno \cdot$ (\Cref{ass:base}),
$\eo := L \sup_{\bx,\by \in  \cX } \lno{ \bx - \by }$, $\e \in (0,\eo)$, $\me := \bce{ \log_2(\eo/\e) }$, $\e_{\me} := \e$, $\e_k := \eo 2^{-k}$ for all $k < \me$, and
\[
	\SC(f,\e)
:= 
    \cN\lrb{ \Xe, \frac{\e}{L} }
    +
	\sum_{k=1}^{\me} 
	\cN\lrb{ \cX_{(\e_k ,\, \e_{k-1}]} ,\, \frac{\e_k}{L} }\;.
\]
If $n \ge \SC(f,\e)$, then the $n$-th error certificate $\xi_n$ of the \CpiyA{} (\Cref{a:cpiya}, run with input $L$, $\lno \cdot$, $\bx_1$) satisfies $\xi_n \le \epsilon$.
\end{theorem}

\begin{proof}
By definition of $\e_k$, we can partition $\cX$ into the $\me+1$ sets
\[
    \cX 
= 
    \cX_\e
    \cup
    \bigcup_{k = 1}^{\me} \cX_{(\e_k ,\, \e_{k-1}]} \;.
\]
Let $\tau$ be the first time where the algorithm returns an error certificate $\xi_\tau \le \e$.
Then, for any $i \in [\tau]$, either $\bx_i \in \Xe$ or there exists a unique $k\in [\me]$ such that $\bx_i \in \cX_{(\e_k ,\, \e_{k-1}]}$.
Since the \CpiyA{} automatically adapts its exploration to all possible approximation scales at the same time, \Cref{l:spread-exploration} and the definition of packing numbers imply that the maximum number of times that a query point $\bx_i$ is chosen in $\Xe$ (up to and including time $\tau$) is upper bounded by 
$
    \cN \lrb{ \Xe ,\, \frac{\e}{L} }
$.
Similarly, since the \CpiyA{} does not explore too much in suboptimal regions, \Cref{l:good-exploration} and the definition of packing numbers imply that for all $k\in [\me]$, the maximum number of times that a query point $\bx_i$ is chosen in $\cX_{(\e_k ,\, \e_{k-1}]} \s \cX_{\e_k}^c$ is upper bounded by 
$
    \cN \lrb{ \cX_{(\e_k ,\, \e_{k-1}]} ,\, \frac{\e_k}{L} }
$.
Hence
\begin{align*}
    \tau
&
=
    \Babs{ \bcb{ i\in[\tau] : \bx_i\in\Xe } }
    +
    \sum_{k=1}^{\me} \Babs{ \bcb{ i\in[\tau] : \bx_i\in \cX_{(\e_k ,\, \e_{k-1}]} } }
\\
&
\le
    \cN \lrb{ \Xe ,\, \frac{\e}{L} }
    +
    \sum_{k=1}^{\me}\cN \lrb{ \cX_{(\e_k ,\, \e_{k-1}]} ,\, \frac{\e_k}{L} }
=
    \SC(f,\e)\;.    
\end{align*}
Therefore, noting that $k\mapsto \xi_k$ is non-increasing (because $\xi_k = \fh_k^\star - f_k^\star$, with $k\mapsto \fh_k^\star$ non-increasing and $k\mapsto f_k^\star$ non-decreasing), if
$
    n
\ge
    \SC(f,\e)
$,
then
$
    n
\ge
    \tau
$,
which in turn yields
$\xi_n \le \xi_\tau \le \e$.
\end{proof}

Though the two bounds in \Cref{t:SNC-upper,t:SC-upper} look very similar, they differ by the term $\cN \lrb{ \cX_{\e} ,\, \frac{\e}{L} }$. This difference is not negligible in general and can be explained with a simple example.
Indeed, consider a constant function $c\colon [0,1]^d \to \R$.
Any non-certified algorithm outputs an $\e$-maximizer of $c$ after $1$ single query at any scale $\e$, because all points in the domain are maxima.
However, the only way to \emph{certify} that the output is $\e$-optimal is essentially to perform a grid-search of $[0,1]^d$ with step-size roughly $\e/L$,
so as to be sure there is no hidden bump of height more than $\e$.
This is reflected in the term $\cN \lrb{ \cX_{\e} ,\, \frac{\e}{L} }$, which is of order $(L/\e)^d$ for constant functions.
At a high level, the more ``constant'' a function is, the easier it is to recommend an $\e$-optimal point, but the harder it is to certify that such recommendation is actually a good recommendation.

\subsection{Optimality}
\label{s:optimality}

We now discuss the optimality of the two bounds (\Cref{t:SNC-upper,t:SC-upper}) we derived in the previous section.

\paragraph*{Non-certified optimality.}

In this paper, we focus on $f$-dependent bounds. 
By its nature, no meaningful lower bounds can be stated in the non-certified setting (because for each given function there is an algorithm that outputs one of its maximizers after only $1$ query). 
For the interested reader, there exist minimax lower bounds over specific classes of functions that match the corresponding upper bounds obtained from $\SNC(f,\e)$, such as when the near-optimality dimension is $d$ \cite{nesterov2003introductory} or $d/2$ \cite{Hor-06-UnknownLipschitz} (see the related works in \Cref{s:relatedworks} for more details).

\paragraph*{Certified optimality.}

One of the interesting features of the certified setting is that it makes it possible to write non-trivial $f$-dependent lower bounds, as shown by \cite{bachoc21-ZerothOrderLipschitzOptimization}.

In this paragraph, we focus on the set $\cF_L$ of globally $L$-\lip{} functions $f$ with $L > \Lip(f)$, where $\Lip(f)$ is the smallest \lip{} constant for $f$.
In other words, we assume that the global \lip{} constant of the unknown objective function is not known exactly.
(We will later briefly discuss the unlikely case in which $\Lip(f)$ is known precisely.)

For any certified algorithm $A$, function $f\in \cF_L$, and accuracy $\e$, we denote by $\sigma(A,f,\epsilon)$ the smallest number of queries $n$ that are needed to get $\xi_n \le \e$, i.e., to find \emph{and certify} an $\e$-maximizer.
The following result holds.

\begin{theorem}[{Bachoc et al.\ \cite[Theorem 2]{bachoc21-ZerothOrderLipschitzOptimization}}] \label{thm:lower:bound:sample:dependent}
For any certified algorithm $A$, globally $L$-\lip{} function $f\in \cF_L$, and accuracy $\e \in (0,\eo]$, it holds that
\[
\sigma(A,f,\epsilon) 
>
\frac{c}{1+m_{\epsilon}}
S_\mathrm{C}(f,\e) \;,
\]
where $\me := \bce{ \log_2(\eo/\e) }$ and $c := 2^{-2} \brb{2^{-7}\brb{1-\Lip(f)/L}}^d$.
\end{theorem}

The previous result states that unless the smallest \lip{} constant of the black-box function $f$ is known exactly, no algorithm can find and certify an $\e$-maximizer with less than $\SC(f,\e)$ queries (up to logarithmic terms).

In the unlikely scenario in which $\Lip(f)$ is known exactly, it is known that $\SC(f,\e)$ remains a tight lower bound (up to logarithmic terms) in dimension $d=1$ \cite[Proposition 2]{bachoc21-ZerothOrderLipschitzOptimization}.
Finding a tight $f$-dependent lower bound for this boundary case in dimension $d\ge 2$ is still an open problem.

\subsection{Integral representation of \texorpdfstring{$\SC(f,\e)$}{Sc(f,eps)}}
\label{s:integralrepresentation}

In this section, we give an integral representation to the quantity $\SC(f,\e)$ we presented in \Cref{t:SC-upper}.
This result was first proven in \cite[Theorem 1]{bachoc21-ZerothOrderLipschitzOptimization}.
We state it and prove it again below, but with tighter multiplicative constants.

The theorem holds under a mild geometrical assumption on $\cX$ that roughly states that the set is ``everywhere $d$-dimensional'', in the sense that all points $\bx \in \cX$ are the center of a sufficiently small ball, a constant portion of which is included in $\cX$.
This natural assumption has already appeared in the past (e.g., \cite{HKM20-SmoothContextualBandits,bachoc21-ZerothOrderLipschitzOptimization}) and is weaker than another classical assumption in the statistics literature (the rolling ball assumption, \cite{cuevas2012statistical,walther1997granulometric}).

\begin{assumption}
\label{ass:geomCx}
Let a norm $\lno \cdot$ on $\R^d$ and a constant $\gamma\in(0,1]$ be both known to the learner. 
For all positive $r < \frac 1 2 \diam(\cX)$,
\[
    \vol\brb{ B_{\lno \cdot }(r,\bx) \cap \cX } 
\ge 
    \gamma \vol\brb{ B_{\lno \cdot }(r,\bzero) } \;.
\]
\end{assumption}
The previous assumption holds for all convex sets with non-empty interior, finite unions of them, and more generally for any set $\cX$ that does not contain isolated points, cusps, or lower dimensional regions (such as segments in $\R^2$).
Not also that $\gamma$ may depend on $d$, e.g., $\gamma = 2^{-d}$ if $\cX$ is a hypercube.

The following intermediate result relates packing numbers with volumes. It first appeared in \cite[Proposition 4]{bachoc21-ZerothOrderLipschitzOptimization}, but we include a slightly updated proof for it in \Cref{s:missing-int-repr}.

\begin{proposition}
\label{p:pack-vol}
Fix any  norm $\lno \cdot$, $\gamma \in (0,1]$, and $L > 0$.
If $\cX \s \R^d$ satisfies \Cref{ass:geomCx}, then, for any function $f\colon \cX \to \R$ that is globally $L$-\lip{} with respect to $\lno \cdot$ and for any $0<w <u < \eo := L \diam (\cX)$, the following inequalities hold:
\[
    \cN\lrb{\cX_u, \frac{u}{L}}
\le
    \frac{1}{\gamma}
    \frac{\vol\brb{ \cX_{(\nicefrac{3}{2})u} } }{ \vol\lrb{ \frac{u}{2L} \ball (1, \bzero) } } \;,
\qquad
    \cN\lrb{\cX_{(w, u]}, \frac{w}{L}}
\le
    \frac{1}{\gamma}
    \frac{\vol\brb{ \cX_{ ( \nicefrac{w}{2} ,\, \nicefrac{3u}{2} ]} } }{ \vol\lrb{ \frac{w}{2L} \ball (1, \bzero) } } \;.
\]
\end{proposition}

The next key lemma shows that sums of volumes of non-overlapping sets of suboptimal points can be lower-bounded by a sum of volumes of overlapping ones.
Compared to the original result in \cite[Lemma 3]{bachoc21-ZerothOrderLipschitzOptimization}, we slightly refine the multiplicative constants.
The proof is deferred to \Cref{s:missing-int-repr}.

\begin{lemma}
\label{l:overlap}
Fix any  norm $\lno \cdot$ and $L > 0$.
Then, for any function $f\colon \cX \to \R$ that is globally $L$-\lip{} with respect to $\lno \cdot$
and for all positive $\e < \eo := L \diam (\cX)$, letting $\me := \bce{ \log_2(\eo/\e) }$, $\e_{\me} := \e$, and $\e_k := \eo 2^{-k}$ for all $k < \me$, the following inequality holds:
\[
    \frac{ \vol \brb{ \cX_{2\e} } }{ \e^d }
    +
    \sum_{k=1}^{\me} \frac{ \vol \brb{ \cX_{( \fracc{\e_k}{2} ,\,  2 \e_{k-1} ]} } }{ \e_{k-1}^d }
\le
    \frac {15} 8 \cdot 4^d \lrb{ \frac{ \vol \brb{ \cX_{\e} } }{ \e^d }
    +
    \sum_{k=1}^{\me} \frac{ \vol \brb{ \cX_{ (\e_k ,\, \e_{k-1} ] } } }{ \e_{k-1}^d } }\;.
\]
\end{lemma}
We can now state our integral characterization of the quantity $\SC(f,\e)$ appearing in \Cref{t:SC-upper}, that controls the number of queries that the \CpiyA{} needs to output a certified $\e$-maximizer.

\begin{theorem}
\label{t:hansen-new}
Fix any  norm $\lno \cdot$, $\gamma \in (0,1]$, and $L > 0$.
Then there exist two positive constants $c,C>0$ depending only on $d, \gamma,L,\lno \cdot$ (e.g., $c := L^d/\vol \brb{ \ball (1,\bzero) }$ and $C := (15/8) (16L)^d/ \gamma \vol \brb{ \ball (1,\bzero) }$) such that:
\begin{enumerate}
    \item For any subset $\cX$ of $\R^d$ that satisfies \Cref{ass:geomCx},
    \item For any function $f\colon \cX \to \R$ that is globally $L$-\lip{} with respect to $\lno \cdot$,
\end{enumerate}
and for all positive $\e < \eo := L \diam (\cX)$, the following inequalities hold:\footnote{We actually prove a stronger result. 
The first inequality holds more generally for any $f$ that is $L$-\lip{} around a maximizer with respect to $\lno \cdot$ and \leb{}-measurable, and does not require $\cX$ to be approximately $d$-dimensional.}
{
\small
\[
    c \lrb{
    \frac{\vol (\cX_\e) }{\e^d}
    +
    \int_{\cX_\e^c} \frac{\mathrm{d}\bx}{\brb{ f(\bxs) - f(\bx) }^d}
    }
\le
    \SC(f,\e)
    \le 
    C \lrb{
    \frac{\vol (\cX_\e) }{\e^d}
    +
    \int_{\cX_\e^c} \frac{\mathrm{d}\bx}{\brb{ f(\bxs) - f(\bx) }^d}
    }
    \;.
\]
}
\end{theorem}

The above result can be restated as $\SC(f,\e) = \Theta \brb{ \int_\cX \frac{\mathrm{d}\bx}{ ( f(\bxs) - f(\bx) + \e )^d} }$ (for a direct proof of this claim, see \Cref{t:hansen} in \Cref{s:missing-int-repr}).
As noted in Section~\ref{s:contributions}, combining this result with \Cref{t:SC-upper} resolves a long-standing conjecture of Hansen et al.\ \cite{hansen1991number}.

\begin{proof}
Fix any $\cX \s \R^d$ and $f\colon \cX \to \R$ satisfying the assumptions of the theorem, and let $\e < \eo$. 
Let also $\me := \bce{ \log_2(\eo/\e) }$, $\e_{\me} := \e$, and $\e_k := \eo 2^{-k}$ for all $k < \me$.
Partition the domain of integration $\cX$ into the following $\me + 1$ sets: 
the set of $\e$-optimal points $\cX_{\e}$ and the $\me$ layers $\cX_{(\e_k ,\, \e_{k-1}]}$, for $k \in [\me]$.

We begin by proving the first inequality.
To this end, for any bounded set $A\s \R^d$ and $r>0$, we consider the $r$-covering number $\cM(A,r)$ of $A$ (with respect to $\lno \cdot$) defined as the smallest cardinality of an $r$-covering of $A$, i.e.,
\[
	\cM(A,r)
:=
	\min \bcb{
		k \in \{1,2,\dots\}  :  \exists \bx_1,\ldots,\bx_k \in \R^d, \forall \bx \in A, \exists i\in [ k ], \lno{ \bx-\bx_i}\le r
	} .
\]
if $A$ is nonempty, zero otherwise.
Then,
\begin{align*}
&
    \frac{\vol (\cX_\e) }{\e^d}
    +
    \int_{\cX_\e^c} \frac{\mathrm{d}\bx}{\brb{ f(\bxs) - f(\bx) }^d}
\le
    \frac{\vol (\cX_\e) }{\e^d}
    + \sum_{k=1}^{\me}\frac{ \vol \brb{ \cX_{(\e_k ,\, \e_{k-1}]} } }{ \e_k^d}
\\
&
\qquad\qquad
\le
    \frac{\cM \brb{ \cX_{\e} ,\, \frac{\e}{L} } \cdot \vol \brb{ \ball (1,\bzero) } \brb{ \frac{\e}{L} }^d }{ \e^d }
\\
&
\qquad\qquad\qquad
    + \sum_{k=1}^{\me}\frac{ \cM \brb{ \cX_{(\e_k ,\, \e_{k-1}]} ,\, \frac{\e_k }{L} } \cdot \vol \brb{ \ball (1,\bzero) } \brb{ \frac{\e_k }{L} }^d }{ \e_k ^d }
\\
&
\qquad\qquad
\le
    \frac{\vol \brb{ \ball (1,\bzero) }}{L^d} \lrb
    {
        \cN \lrb{ \cX_{\e} ,\, \frac{\e}{L} } + 
        \sum_{k=1}^{\me} \cN \lrb{ \cX_{(\e_k ,\, \e_{k-1}]} ,\, \frac{\e_k }{L} }
    }
\\
&
\qquad\qquad
=
    \frac{\vol \brb{ \ball (1,\bzero) }}{L^d} 
    \SC(f,\e)
    \;,
\end{align*}
where the first inequality follows by lower bounding $f(\bxs) - f$ with its infimum on the partition, the second one by upper bounding the volume of a set with the volume of the balls of a smallest $\nicefrac{\e_k}{L}$-covering, and the last one by the fact that covering numbers are always upper bounded by packing numbers (we recall this known result in \Cref{sec:lemmaPacking}, \cref{eq:wainwright}).
This proves the first part of the theorem.

For the second one, we have
\begin{align*}
&
    \frac{\vol (\cX_\e) }{\e^d}
    +
    \int_{\cX_\e^c} \frac{\mathrm{d}\bx}{\brb{ f(\bxs) - f(\bx) }^d}
\ge
    \frac{\vol (\cX_{\e}) }{\e^d}
    + \sum_{k=1}^{\me}\frac{ \vol \brb{ \cX_{(\e_k, \e_{k-1}]} } }{ \e_{k-1}^d}
\\
&
\qquad\qquad
\ge
    \frac 8 {15} \frac{1}{4^d} \lrb{
        \frac{ \vol \brb{ \cX_{2\e}} }{ \e^d }
        +
        \sum_{k=1}^m \frac{ \vol \brb{ \cX_{( \fracc{\e_k}{2}, \,  2 \e_{k-1} ]} } }{ \e_{k-1}^d }
    }
\\
&
\qquad\qquad
\ge
    \frac 8 {15}
    \frac
    { 
        \gamma \vol\lrb{ \frac{\e}{2L} \ball (1, \bzero) } \cN\lrb{\cX_{\e}, \frac{\e}{L}}
    }
    {
        (4 \, \e)^d
    }
\\
&
\qquad\qquad\qquad
    +
    \frac 8 {15}
    \sum_{k=1}^{\me} \frac
    { 
        \gamma \vol\lrb{ \frac{\e_k}{2L} \ball (1, \bzero) } \cN\lrb{\cX_{(\e_k, \e_{k-1}]}, \frac{\e_k}{L}} 
    }
    { 
        (4 \, \e_{k-1} )^d
    }
\\
&
\qquad\qquad
\ge
    \frac 8 {15} \frac{\gamma \vol \brb{ \ball (1, \bzero) }}{(8L)^d}
    \, \cN\lrb{\cX_{\e}, \frac{\e}{L}}
\\
&
\qquad\qquad\qquad
    + 
    \frac 8 {15} \frac{\gamma \vol \brb{ \ball (1, \bzero) }}{(16L)^d}
    \sum_{k=1}^{\me} 
    \cN\lrb{\cX_{(\e_k, \e_{k-1}]}, \frac{\e_k}{L}}
    \;,
\end{align*}
where the first inequality follows by upper bounding $f(\bxs) - f$ with its supremum on the partition, the second one by lower bounding the sum of disjoint layers with that of overlapping ones (\Cref{l:overlap}), the third one by the elementary inclusions 
$\cX_{2 \e } \supseteq \cX_{\frac{3}{2} \e}$
and
$\cX_{\left(\frac{1}{2} \e_k ,\, 2\e_{k-1}\right]} \supseteq \cX_{\left(\frac{1}{2} \e_k ,\, \frac{3}{2} \e_{k-1}\right]}$ (for all $k \in [\me]$) followed by a relationship between packing numbers and volumes (\Cref{p:pack-vol}) that holds under \Cref{ass:geomCx}, and the last one by $\e_{k-1}\le 2\e_k$ (for all $k \in [\me]$).
\end{proof}

\section{Robustness}
\label{s:robustness}

In this section, we consider a harder setting in which, for each $k\ge 1$, querying the $k$-th value of $f$ at a point $\bx_k$ reveals a perturbed evaluation $f(\bx_k) + \alpha_k$, where $\alpha_k \in [-\alpha,\alpha]$ is the $k$-th element of an adversarial sequence of perturbations.
Both the sequence $(\alpha_k)_{k\ge 1}$ and $\alpha\ge 0$ are assumed to be unknown in the non-certified setting, while in order to output a certificate, $\alpha$ has to be known in the certified setting.
Furthermore, we now only assume that $\cX$ is bounded, lifting the assumption that $\cX$ is closed.

\subsection{\apiya{}}

In this section, we present a natural extension of the \piya{} to our setting with perturbations and a bounded domain.
\Cref{a:acpiya} proceeds similarly to \Cref{a:cpiya}, but it only approximately maximizes the proxy function after each update.
This is not only necessary when $\cX$ is not closed, but it could prove useful in general whenever the task of optimizing $\fh_k$ is computationally expensive, which, as discussed in \Cref{s:defPS}, is often the case (except in small dimensions).

\begin{algorithm}
\caption{Approximate \piy{}, for the non-certified (/\emph{certified}) setting}
\label{a:acpiya}
\textbf{input:} \lip{} constant $L>0$, norm $\lno{\cdot}$, initial guess $\bx_1 \in \cX$, tolerance $\eta\ge 0$ (\emph{and magnitude of the perturbations $\alpha\ge 0$, certified setting only})
\begin{algorithmic}
\For{$k = 1, 2, \dots$}
    \State observe the perturbed value $g(\bx_k) \gets f(\bx_k) + \alpha_k$
    \State output the recommendation $\bxs_k \gets \argmax_{\bx\in\{\bx_1, \ldots, \bx_k\}} g(\bx) $
    \State update the proxy function
    \[
        \widehat f_k(\cdot) \gets \min_{i\in[k]} \Bcb{ g(\bx_i) + L \bno{ \bx_i - (\cdot) } }
    \]
    \State pick as the next query point any $\bx_{k+1} \in \cX$  such that $\fh_k(\bx_{k+1}) \ge \sup_{\bx\in  \cX } \fh_k(\bx) - \eta$
    \State output the error certificate $\xi_k \gets \widehat f^\star_{k} - f^\star_{k} + (2\alpha+\eta)$, where
    \[
        \widehat f ^\star_k \gets \fh_k(\bx_{k+1}) \;, \quad
        f^\star_k \gets g(\bxs_k)
        \quad \text{(\emph{certified setting only})}
    \]
\EndFor
\end{algorithmic}
\end{algorithm}

\subsection{Upper bounds}
\label{s:upperbounds-alpha}

In this section, we show that if $\alpha,\eta \ll 1$, the number of queries that the \apiya{} needs to reach (or certify) a given accuracy are comparable to those of the \piya{} in the simpler setting without perturbations.
In other words, the paradigm imposed by the \piya{} is robust to small perturbations or approximation errors.
The proofs of the results of this section require more technical work but follow the same high level ideas of the corresponding ones in \Cref{s:lemmas,s:upperbounds}.
For this reason, we defer all of them to \Cref{s:missing-upperbounds}.

We begin by stating an upper bound on the number of evaluations of $f$ that are needed before the recommendations of the \apiya{} become near-optimal.

\begin{theorem}
\label{t:SNC-upper-alpha}
Assume that $f\colon \cX \to \R$ is $L$-\lip{} around a maximizer $\bxs$ with respect to a norm $\lno \cdot$ (\Cref{ass:base}),
queries are corrupted by additive perturbations with absolute value bounded by some unknown $\alpha\ge 0$,
$\eta \ge 0$, $\eo := L \sup_{\bx,\by \in  \cX } \lno{ \bx - \by }$, $\e \in (0,\eo)$, $\me := \bce{ \log_2(\eo/\e) }$, $\e_{\me}^{\alpha,\eta} := \e + 2\alpha+\eta$, $\e_k^{\alpha,\eta} := \eo 2^{-k} + 2\alpha+\eta$ for all $k < \me$, and
\[
	\SNC^{\alpha,\eta}(f,\e)
:= 
	\sum_{k=1}^{\me} 
	\cN\lrb{ \cX_{(\e_k^{\alpha,\eta} ,\, \e_{k-1}^{\alpha,\eta} ]} ,\, \frac{\e_k^{\alpha,\eta} - 2\alpha-\eta }{L} }\;.
\]
If $n \ge 1+\SNC^{\alpha,\eta}(f,\e)$, then the $n$-th recommendation $\bxs_n$ of the \apiya{} (\Cref{a:acpiya}, run with input $L$, $\lno \cdot$, $\bx_1$, $\eta$) satisfies $f(\bxs)-f(\bxs_n) \le \epsilon + 4\alpha+\eta$.
\end{theorem}

Similarly to the previous result, we now generalize \Cref{t:SC-upper} showing that when the evaluations of $f$ are perturbed by an adversarial corruption of magnitude at most $\alpha$, we can control the number of queries $n$ that the \AcpiyA{} needs to guarantee that the error certificate $\xi_n$ is smaller than $\e + 2\alpha+\eta$.

\begin{theorem}
\label{t:SC-upper-alpha}
Assume that $f\colon \cX \to \R$ is $L$-\lip{} around a maximizer $\bxs$ with respect to a norm $\lno \cdot$ (\Cref{ass:base}),
queries are corrupted by additive perturbations with absolute value bounded by some known $\alpha\ge 0$,
$\eta \ge 0$, $\eo := L \sup_{\bx,\by \in  \cX } \lno{ \bx - \by }$, $\e \in (0,\eo)$, $\me := \bce{ \log_2(\eo/\e) }$, $\e_{\me}^{\alpha,\eta} := \e + (2\alpha+\eta)$, $\e_k^{\alpha,\eta} := \eo 2^{-k} + (2\alpha+\eta)$ for all $k < \me$, and
\[
	\SC^{\alpha,\eta}(f,\e)
:= 
    \cN\lrb{ \cX_{\e_{\me}^{\alpha,\eta}} ,\, \frac{\e_{\me}^{\alpha,\eta}-2\alpha}{L} }
    +
	\sum_{k=1}^{\me} 
	\cN\lrb{ \cX_{(\e_k^{\alpha,\eta} ,\, \e_{k-1}^{\alpha,\eta} ]} ,\, \frac{\e_k^{\alpha,\eta} - 2\alpha-\eta }{L} }\;.
\]
If $n \ge \SC^{\alpha,\eta}(f,\e)$, then the $n$-th error certificate $\xi_n$ of the \AcpiyA{} (\Cref{a:acpiya}, run with input $L$, $\lno \cdot$, $\bx_1$, $\eta$, $\alpha$) satisfies $\xi_n \le \epsilon + 2\alpha+\eta$.
\end{theorem}

\section{Related Works}
\label{s:relatedworks}

We now discuss related works, both about the \piya{} itself and close variants from the global optimization and bandit optimization literature.

\paragraph*{Known bounds for the \piya{} (non-certified setting).} 
Several papers studied the \piya{} in the eighties and nineties (see, e.g., \cite{mayne1984outer,mladineo1986algorithm,hansen1991number} or the survey by \cite{HansenETAL-92-Survey}). Despite this literature, little was known about the rate of convergence of its optimization error $f(\bxs)-f(\bxs_n)$ as a function of the number $n$ of evaluations of $f$. A crude bound of the form $f(\bxs)-f(\bxs_n) = \mathcal{O}\bigl(n^{-1/d}\bigr)$, where $d$ is the ambient dimension, can be obtained from \cite[Theorem~4.2]{mladineo1986algorithm} when $f$ is globally Lipschitz. The authors showed that the \piya{} is minimax optimal among all algorithms, and therefore superior to a uniform grid search ensuring an optimization error of the order of $n^{-1/d}$. 

\paragraph*{Known bounds for the \piya{} (certified setting).}
In dimension $d=1$ an $f$-dependent bound on the sample complexity was derived by Hansen et al.\ \cite{hansen1991number} for the certified version of the \piya{}. Rephrasing their main result into our own terms, they proved that $\epsilon$-accuracy is guaranteed (that is, $\xi_n \le \epsilon$) after a number of iterations at most proportional to $\int_0^1 \brb{ f(\xs)-f(x)+\e }^{-1} \dif x$.
The authors relied heavily on the one-dimensional setting to study the proxy functions $\fh_k$ in an explicit manner, and claimed that ``\emph{Extending the results of this paper to the multivariate case appears to be difficult}''. To the best of our knowledge, we provide the first such bound in higher dimensions $d \ge 2$.

\paragraph*{Other packing-based bounds (with other non-certified bandit algorithms).} 
The non-certified version of our global Lipschitz optimization problem has been well studied in the bandit optimization literature over the past decades; see the textbooks by Munos \cite{munos2014bandits} and Slivkins \cite{slivkins2019introduction}. Several algorithms have been developed, with various $f$-dependent error bounds involving sets of $\e$-optimal points $\cX_\e$ or layers $\cX_{(a,b]}$ of $f$. For instance, for the DOO algorithm, which can be seen as a discretized version  of the \piya{}, Munos \cite[Theorem~1]{munos2011optimistic} proved a bound roughly of the form $\sum_{i=1}^{\delta^{-1}(\e)} \cN(\cX_{\delta(i)},c \delta(i))$ for some constant $c>0$ independent of $f$, and where the sequence $i \mapsto \delta(i)$ is decreasing. In the special case where $\delta(i) \approx \gamma^{i}$ with $\gamma \in (0,1)$, and for (weakly) Lipschitz functions $f$ such that $\cN(\cX_{\e},c\e) \lesssim (1/\e)^{d^\star}$ with $d^\star \in [0,d]$ (we say that $f$ has near-optimality dimension smaller than $d^\star$), this sample complexity bound implies a bound roughly of $\log(1/\e)$ if $d^\star = 0$ or $(1/\e)^{d^\star}$ if $d^\star >0$. A similar bound was proved by Perevozchikov \cite{Per-90-OptimizationComplexity} for a branch-and-bound algorithm close to the DOO algorithm, with an assumption on the volume of $\cX_\e$ (instead of a packing number), or by Malherbe and Vayatis \cite{MaVa-17-LipBandits} for a stochastic version of the Piyavskii-Shubert algorithm under a stronger assumption on the shape of $f$ around its maximizer. Rather than maximizing the proxy function $\fh_{k-1}$, the latter algorithm (called LIPO) queries the $k$-th point uniformly at random in the set of potential maximizers $\bcb{x\in \cX : \fh_{k-1}(x) \ge \max_{i=1,\ldots,k-1} f(x_i) }$. 

The matching worst-case lower bounds of  Nesterov \cite[Theorem~1.1.2]{nesterov2003introductory} when $d^\star=d$, of Horn \cite{Hor-06-UnknownLipschitz} when $d^\star = d/2$, and the lower bounds of Bubeck \cite{bubeck2011x} for any $d^\star$ but in the stochastic case suggest that this bound is worst-case optimal over the class of functions with given near-optimality dimension $d^\star$. However, the bounds mentioned above have several limitations. First, as noted earlier by Kleinberg et al.\ \cite[remark after Theorem~4.4]{kleinberg2019bandits}, bounds involving a single notion of dimension such as the near-optimality or zooming dimension might be too crude. Indeed, functions that feature different shapes at different scales (such as, e.g., different values of $d^\star$ for different ranges of $\epsilon$) are better analyzed in a non-asymptotic framework through sums of packing numbers at different scales. Second, considering packing numbers of the near-optimal sets $\cX_{\delta(i)}$ as in the first bound mentioned above can also be quite suboptimal. For instance, for constant functions $f$, the sets $\cX_{\delta(i)}$ are equal to the whole domain $\cX$ (maximal packing number) while $f$ is optimized by any algorithm after only a single evaluation of~$f$. The second limitation was overcome in general metric spaces by Kleinberg et al.\ \cite{kleinberg2008multi,kleinberg2019bandits} by considering packing numbers of layers $\cX_{(a,b]}$ instead of near-optimal sets $\cX_{\e}$.

Our analysis of the \piya{} indicates that this very natural algorithm can also handle multiple shapes at different scales, with a sample complexity bound also involving layers of $f$, and that this also holds in the certified setting.

\paragraph*{Relationship with Bachoc et al.\ \cite{bachoc21-ZerothOrderLipschitzOptimization}.} The very recent paper \cite{bachoc21-ZerothOrderLipschitzOptimization}, about optimal instance-dependent sample complexity bounds in the certified setting, was mainly written after but published before the present paper. Taking inspiration from our $S_C$ bound and analysis, the authors prove a similar bound for a certified version of the more practical DOO algorithm. They also provide a nearly matching instance-dependent lower bound, as well as an equivalent integral representation of $S_C$. For the sake of completeness, we recalled these two complementary results in Sections~\ref{s:optimality} and~\ref{s:integralrepresentation}, while improving the multiplicative constants in the integral representation.

\paragraph*{Adaptivity to $L$.}

In the context of global optimization, there is a line of research devoted to maximizing $L$-Lipschitz functions \emph{without} knowing $L$. 
Algorithms for these problems typically sacrifice some of the available evaluations of $f$ to compute an estimate of the Lipschitz constant $L$ and use the remaining ones to optimize the objective. 
Such estimates $\widehat{L}$ of $L$ can be defined (up to small margins) as $\widehat{L} = \max_{i\neq j} \frac{f(x_i)-f(x_j)}{x_i-x_j}$, where the maximum is over all pairs of distinct values $x_i, x_j$ that have been queried so far.  
Bubeck et al.\ \cite{bubeck2011lipschitz} present a two-phase algorithm in which the estimation of the Lipschitz constant occurs in the first phase. Their results hold for twice differentiable objectives for which the eigenvalues of the Hessian are bounded everywhere by a constant $M$. 
Notably, the bounds they obtain are meaningless unless the time horizon is bigger than a function of $M$ (i.e., in order to apply the result in practice one would need prior knowledge of $M$).

Malherbe and Vayatis \cite{MaVa-17-LipBandits} introduce AdaLIPO, an adaptive version of LIPO (see above).
Their algorithm flips a biased coin at each round. Depending on the outcome, AdaLIPO either performs a LIPO step (using the current best estimate of the Lipschitz constant) or it evaluates the function at a point drawn uniformly in the domain of $f$. These uniform draws are used to obtain increasingly more accurate approximations of the Lipschitz constant.
Again, the bounds they obtain become meaningless if the probability of getting a good estimate of the Lipschitz constant with only two uniform queries to $f$ (what the authors call $\Gamma$, in Proposition~18) is too small.
In some papers the Lipschitzness assumption is replaced by some other regularity assumption (e.g., \cite{bartlett2019simple,grill2015black} use a notion of smoothness related to hierarchical partitioning that is comparable to our Assumption~\ref{ass:base}).

The idea of investing some of the available evaluations to approximate the Lipschitz constant could also be applied to the \piya{} in the non-certified setting. 
However, missing any prior quantitative information on the smoothness of $f$ has a crucial limitation. 
It makes it \emph{impossible} to design algorithms that certify the accuracy of their recommendations (without any other inputs). 
To see this, note that \emph{all} algorithms will fail when the objective function is the ``spike'' $x \mapsto \max \bcb{0, a - L \| x - x_0\|}$ for some values of $a$, $L$, and $x_0$ that depend on the algorithm (typically $a$ and $L$ should be chosen large enough to force the algorithm to miss a tall and narrow spike).

\section*{Acknowledgements}

The authors would like to thank Jean-Baptiste Hiriart-Urruty, Marcel Mongeau, Edouard Pauwels, and Michal Valko for their helpful discussions and bibliographic insights. 

\section*{Funding}

S\'{e}bastien Gerchinovitz gratefully acknowledges the support of the DEEL project.\footnote{\url{https://www.deel.ai/}}
Tommaso Cesari gratefully acknowledges the support of the European Network for Game Theory (GAMENET) and the project BOLD from the French national research agency (ANR). 
This project has also received funding from the French ``Investing for the Future – PIA3'' program under the Grant agreement ANITI ANR-19-P3IA-0004.

\bibliographystyle{tfs}
\bibliography{biblio}


\renewcommand{\theHsection}{A\arabic{section}}

\appendix

\section{Standard inequalities about packing and covering numbers}
\label{sec:lemmaPacking}

We recall that for any norm $\lno \cdot$, bounded set $A\s\R^d$, and $r>0$: 
\begin{itemize}
    \item The $r$-\emph{packing number} of $A$ (with respect to $\lno \cdot$) is the largest number of $r$-separated points contained in $A$, i.e.,
    \[
    	\cN(A,r)
    := 
    	\sup \lcb{
    		k \ge 1  :  \exists \bx_1, \dots, \bx_k \in A, \min_{i\neq j} \lno{ \bx_i - \bx_j } > r
    	} \;,
    \]
    if $A$ is nonempty, zero otherwise
    \item The \emph{$r$-covering number} of $A$ (with respect to $\lno \cdot$) is the smallest number of points needed to cover $A$ with balls with radius $r$ centered at those points, i.e.,
    \[
    	\cM(A,r)
    :=
    	\min \bcb{
    		k \ge 1  :  \exists \bx_1,\ldots,\bx_k \in \R^d, \forall \bx \in A, \exists i\in [ k ], \lno{ \bx-\bx_i}\le r
    	} .
    \]
    if $A$ is nonempty, zero otherwise
\end{itemize}
Covering and packing numbers are closely related. In particular, the following well-known inequalities hold---see, e.g., \citep[Lemmas~5.5~and~5.7, with permuted notations for $\cM$ and $\cN$]{Wainwright19-HighDimensionalStatistics}.\footnote{The definition of $r$-covering number of a subset $A$ of $\R^d$ implied by \citep[Definition 5.1]{Wainwright19-HighDimensionalStatistics} is slightly stronger than the one used in our paper, because elements $x_1, \ldots, x_N$ of $r$-covers belong to $A$ rather than just $\R^d$.}
\begin{lemma}
For any bounded set $A\s\R^d$ and any real number $r>0$,
\begin{equation}
\label{eq:wainwright}
	\cN(A,2r)
\le
	\cM(A,r)
\le
	\cN(A,r)\;.
\end{equation}
\end{lemma}

\section{Missing proofs of Section \ref{s:integralrepresentation}}
\label{s:missing-int-repr}

In this section, we include all missing details from \Cref{s:integralrepresentation}.

\begin{proof}[Proof of Proposition \ref{p:pack-vol}.]
Fix any $0<w<u<\eo$. 
Let $\eta_1:=\frac{u}{L}$, $\eta_2:=\frac{w}{L}$, $E_1:=\cX$, $E_2:=\cX_w^c$, and $i\in[2]$.
Note that for any $\eta>0$ and $A \s \cX$, the balls of radius $\nicefrac{\eta}{2}$ centered at the elements of an $\eta$-packing of $A$ intersected with $\cX$ are all disjoint and included in $\brb{ A+B_{\lno \cdot}(\nicefrac{\eta}{2}, \bzero) } \cap \cX$ (see \Cref{f:illuminati}).
\begin{figure}
    \centering
    \begin{tikzpicture}
    \draw[dashed] ({-3-sqrt(2)},-1) -- ({3+sqrt(2)},-1) -- (0,{2+sqrt(2)}) -- cycle;
    \begin{scope}
        \clip (-2.5,-0.5) rectangle (2.5, 2.5);
        \fill[purple!30!blue!50!white] (0,0)  circle (1)
          (-2,0) circle (1)
          (2,0)  circle (1)
          (0,2) circle (1);
        \end{scope}
    \draw (0,0)  circle (1)
          (-2,0) circle (1)
          (2,0)  circle (1)
          (0,2)  circle (1);
    \fill (0,0)  circle (1.5pt)
          (-2,0) circle (1.5pt)
          (2,0)  circle (1.5pt)
          (0,2)  circle (1.5pt);
    \draw (-2,0) -- (2,0) -- (0,2) -- cycle;
    \draw (-2.5,-0.5) rectangle (2.5, 2.5);
    \draw[very thick] (-2.5, -0.5) 
            -- (-2.5, {-0.5+sqrt(2)}) 
            -- ({0.5-sqrt(2)}, 2.5)
            -- ({-0.5+sqrt(2)}, 2.5)
            -- (2.5, {-0.5+sqrt(2)}) 
            -- (2.5, -0.5)
            -- cycle;
    \end{tikzpicture}
    \caption{An illustration that that for any $\eta>0$ (the radii of the circles) and $A \s \cX$ (inner triangle and rectangle respectively), the balls of radius $\nicefrac{\eta}{2}$ centered at the elements of an $\eta$-packing of $A$ (the circles centered at the dots on the perimeter of $A$) intersected with $\cX$ are all disjoint and included in $\brb{ A+B_{\lno \cdot}(\nicefrac{\eta}{2}, \bzero) } \cap \cX$ (the outer dashed triangle intersected with the rectangle, which give the bold hexagon containing the colored figures).}
    \label{f:illuminati}
\end{figure}
Thus, letting $P_i$ be a set of $\eta_i$-separated points included in $A_i := \cX_u\cap E_i$ with cardinality $\labs {P_i} = \cN(A_i, \eta_i)$, we have
\begin{align*}
    \vol\Brb{ \brb{ A_i + \ball ( \nicefrac{\eta_i}{2}, \bzero ) } \cap \cX }
&
\ge
    \sum_{\bx \in P_i}
    \vol\brb{ \ball (\nicefrac{\eta_i}{2},\bx) \cap \cX }
\\
&
\ge
    \gamma \vol\brb{ \ball (\nicefrac{\eta_i}{2},\bzero) } \cN\lrb{ A_i , \eta_i }
    \;,
\end{align*}
where the second inequality follows by Assumption~\ref{ass:geomCx}.
We now further upper bound the left-hand side. 
Take an arbitrary point $\bx_i \in \brb{ \cX_u \cap E_i + \ball ( \nicefrac{\eta_i}{2}, \bzero ) } \cap \cX$. 
By definition of \mink{} sum, there exists $\bx_i' \in \cX_u \cap E_i$ such that $\lno{\bx_i - \bx'_i } \le \nicefrac{\eta_i}{2}$.
Hence 
$
    f(\bxs) - f(\bx_i) 
\le 
    f(\bxs) - f(\bx_i') + \babs{ f(\bx_i') - f(\bx_i) } 
\le 
    u + L (\nicefrac{\eta_i}{2}) 
\le 
    (\nicefrac{3}{2}) u$.
This implies that $\bx_i \in \cX_{(\nicefrac{3}{2})u}$, which proves the first inequality. 
For the second one, note that $\bx_2$ satisfies 
$
    f(\bxs) - f(\bx_2) 
\ge 
    f(\bxs) - f(\bx_2') - \babs{ f(\bx_2') - f(\bx_2) } 
\ge 
    w - L (\nicefrac{\eta_2}{2}) 
=
    (\nicefrac{1}{2}) w$.
\end{proof}

\begin{proof}[Proof of Lemma \ref{l:overlap}.]
Fix any $\cX$, $f$, and $\e$ as in the statement of the lemma.
To avoid clutter, we denote $\me$ simply by $m$.

Assume first that $m\ge 3$. 
Then
\begin{align*}
& 
\frac{ \vol \brb{ \cX_{2\e}} }{ \e^d }
    +
    \sum_{k=1}^m \frac{ \vol \brb{ \cX_{( \fracc{\e_k}{2}, \,  2 \e_{k-1} ]} } }{ \e_{k-1}^d }
\le
    \frac
    { 
        \vol \brb{ \cX_{\e} }
        +
        \vol \brb{ \cX_{ ( \e_m, \, \e_{m-1} ] } }
        +
        \vol \brb{ \cX_{ ( \e_{m-1}, \, \e_{m-2} ] } } 
    }{ \e^d }
\\
& 
\qquad\qquad\qquad
+
    \sum_{k=1}^{m-2}\frac{\vol\brb{\cX_{\left(\e_{k+1},\,\e_{k}\right]}} + \vol\brb{\cX_{\left(\e_{k},\,\e_{k-1}\right]}} + \vol\brb{\cX_{\left(\e_{k-1},\,\e_{k-2}\right]}}}{\e_{k-1}^{d}}
\\
& 
\qquad\qquad\qquad
+
    \frac
    {
        \vol \brb{ \cX_{\e} }
        +
        \vol \brb{ \cX_{ ( \e_m, \, \e_{m-1} ] } }
        +
        \vol \brb{ \cX_{ ( \e_{m-1}, \, \e_{m-2} ] } }
        +
        \vol \brb{ \cX_{ ( \e_{m-2}, \, \e_{m-3} ] } }
    }
    { \e_{m-2}^d }
\\
&
\qquad\qquad\qquad
+
    \frac{
        \vol \brb{ \cX_{\e} }
        +
        \vol \brb{ \cX_{ ( \e_m, \, \e_{m-1} ] } }
        +
        \vol \brb{ \cX_{ ( \e_{m-1}, \, \e_{m-2} ] } }
    }{ \e_{m-1}^d }
\\
& 
\qquad
\le
    \lrb{ 2 + \frac 1 {2^d} }\frac{\vol\brb{\cX_{\e}}}{\e^{d}}
    +
    \lrb{ 2^d + 1 + \frac 1 {2^d} }\frac{\vol\brb{\cX_{(\e_{m}, \,\e_{m-1}]}}}{\e_{m-1}^{d}}
\\
&
\qquad\qquad\qquad
    +
    \brb{ 4^{d}+2^{d}+1 } \frac{\vol\brb{\cX_{(\e_{m-1}, \,\e_{m-2}]}}}{\e_{m-2}^{d}}
\\
& 
\qquad\qquad\qquad
+
\frac{1}{2^{d}}\sum_{k=2}^{m-1}\frac{\vol\brb{\cX_{\left(\e_{k},\,\e_{k-1}\right]}}}{\e_{k-1}^{d}}+\sum_{k=1}^{m-2}\frac{\vol\brb{\cX_{\left(\e_{k},\,\e_{k-1}\right]}}}{\e_{k-1}^{d}}+2^{d}\sum_{k=1}^{m-2}\frac{\vol\brb{\cX_{\left(\e_{k},\,\e_{k-1}\right]}}}{\e_{k-1}^{d}}
\\
& 
\qquad
=
    \lrb{ 2 + \frac 1 {2^d} } \frac{\vol\brb{\cX_{\e}}}{\e^{d}}
    +
    4^{d}\frac{\vol\brb{\cX_{(\e_{m-1}, \,\e_{m-2}]}}}{\e_{m-2}^{d}}
\\
&
\qquad\qquad\qquad
+
    \frac{1}{2^{d}}\sum_{k=2}^{m}\frac{\vol\brb{\cX_{\left(\e_{k},\,\e_{k-1}\right]}}}{\e_{k-1}^{d}}
    +
    \sum_{k=1}^{m}\frac{\vol\brb{\cX_{\left(\e_{k},\,\e_{k-1}\right]}}}{\e_{k-1}^{d}}
    +
    2^{d}\sum_{k=1}^{m}\frac{\vol\brb{\cX_{\left(\e_{k},\,\e_{k-1}\right]}}}{\e_{k-1}^{d}}
\\
&
\qquad
\le
    \lrb{ 2 + \frac 1 {2^d} } 
    \frac{ \vol \brb{ \cX_{\e} } }{ \e^d }
    +
    \lrb{ 4^d + \frac 1 {2^d} + 1 + 2^d }
    \sum_{k=1}^m \frac{ \vol \brb{ \cX_{ (\e_k, \, \e_{k-1}] } } }{\e_{k-1}^{d}}
\end{align*}
where we applied several times the definitions of and the relations between the $\e_{k}$'s.
The bound then follows after noting that $2+2^{-d} \le 4^d + 2^{-d} + 1 + 2^d \le (15/8) \cdot 4^d$.

Assume now that $m=2$. 
Then
\begin{align*}
&
    \frac{ \vol \brb{ \cX_{2\e}} }{ \e^d }
    +
    \sum_{k=1}^m \frac{ \vol \brb{ \cX_{( \fracc{\e_k}{2}, \,  2 \e_{k-1} ]} } }{ \e_{k-1}^d }
\\
&
\qquad
\le
    \frac
    { 
        \vol \brb{ \cX_{\e} }
        +
        \vol \brb{ \cX_{ ( \e_m, \, \e_{m-1} ] } }
        +
        \vol \brb{ \cX_{ ( \e_{m-1}, \, \e_{m-2} ] } } 
    }{ \e^d }
\\
& 
\qquad\qquad
+
    \frac
    {
        \vol \brb{ \cX_{\e} }
        +
        \vol \brb{ \cX_{ ( \e_m, \, \e_{m-1} ] } }
        +
        \vol \brb{ \cX_{ ( \e_{m-1}, \, \e_{m-2} ] } }
    }
    { \e_{m-2}^d }
\\
&
\qquad\qquad
+
    \frac{
        \vol \brb{ \cX_{\e} }
        +
        \vol \brb{ \cX_{ ( \e_m, \, \e_{m-1} ] } }
        +
        \vol \brb{ \cX_{ ( \e_{m-1}, \, \e_{m-2} ] } }
    }{ \e_{m-1}^d }
\\
& 
\qquad
\le
    \lrb{ 2 + \frac 1 {2^d} } \frac{ \vol \brb{ \cX_{\e}} }{ \e^d }
    +
    \lrb{ 4^d + 1 + 2^d }
    \sum_{k=1}^m \frac{ \vol \brb{ \cX_{ (\e_k, \, \e_{k-1}] } } }{\e_{k-1}^{d}}
\end{align*}
and the bound follows from $2+2^{-d} \le 4^d + 1 + 2^d \le (7/4) \cdot 4^d \le (15/8) \cdot 4^d$.

Finally, assume that $m=1$. 
Then
\begin{align*}
    \frac{ \vol \brb{ \cX_{2\e}} }{ \e^d }
    +
    \sum_{k=1}^m \frac{ \vol \brb{ \cX_{( \fracc{\e_k}{2}, \,  2 \e_{k-1} ]} } }{ \e_{k-1}^d }
& \le
    \frac
    { 
        \vol \brb{ \cX_{\e} }
        +
        \vol \brb{ \cX_{ ( \e_m, \, \e_{m-1} ] } }
    }{ \e^d }
\\
&
\qquad
    +
    \frac{
        \vol \brb{ \cX_{\e} }
        +
        \vol \brb{ \cX_{ ( \e_m, \, \e_{m-1} ] } }
    }{ \e_{m-1}^d }
\\
& \le
    2 \frac{ \vol \brb{ \cX_{\e}} }{ \e^d }
    +
    2^d
    \sum_{k=1}^m \frac{ \vol \brb{ \cX_{ (\e_k, \, \e_{k-1}] } } }{\e_{k-1}^{d}}
\end{align*}
and the bound follows from $2 \le 2^d \le (15/8) \cdot 4^d$.
This concludes the proof.
\end{proof}

\begin{theorem}
\label{t:hansen}
Fix any  norm $\lno \cdot$, $\gamma \in (0,1]$, and $L > 0$.
Then there exist two positive constants $c,C>0$ depending only on $d, \gamma,L,\lno \cdot$ (e.g., $c := L^d/\vol \brb{ \ball (1,\bzero) }$ and $C := (15/8) (32L)^d/ \gamma \vol \brb{ \ball (1,\bzero) }$) such that:
\begin{enumerate}
    \item For any subset $\cX$ of $\R^d$ that is compact and $\gamma$-approximately $d$-dimensional with respect to $\lno \cdot$,
    \item For any function $f\colon \cX \to \R$ that is globally $L$-\lip{} with respect to $\lno \cdot$,
\end{enumerate}
and for all positive $\e < \eo := L \diam (\cX)$, the following inequalities hold:\footnote{We actually prove a stronger result. 
The first inequality holds more generally for any $f$ that is $L$-\lip{} around a maximizer with respect to $\lno \cdot$ and \leb{}-measurable, and does not require $\cX$ to be approximately $d$-dimensional.}
\[
    c
    \int_\cX \frac{\mathrm{d}\bx}{\brb{ f(\bxs) - f(\bx) + \e }^d}
\le
    \SC(f,\e)
    \le 
    C
    \int_\cX \frac{\mathrm{d}\bx}{\brb{ f(\bxs) - f(\bx) + \e }^d}
    \;.
\]
\end{theorem}

\begin{proof}
Fix any $\cX \s \R^d$ and $f\colon \cX \to \R$ satisfying the assumptions of the theorem, and let $\e < \eo$. 
Let also $\me := \bce{ \log_2(\eo/\e) }$, $\e_{\me} := \e$, and $\e_k := \eo 2^{-k}$ for all $k < \me$.
Partition the domain of integration $\cX$ into the following $\me + 1$ sets: 
the set of $\e$-optimal points $\cX_{\e}$ and the $\me$ layers $\cX_{(\e_k, \e_{k-1}]}$, for $k \in [\me]$.

We begin by proving the first inequality.
To this end, for any bounded $A\s \R^d$ and $r>0$, we define the $r$-covering number $\cM(A,r)$ of $A$ (with respect to $\lno \cdot$) as the smallest cardinality of an $r$-covering of $A$, i.e.,
\[
	\cM(A,r)
:=
	\min \bcb{
		k \in \{1,2,\dots\}  :  \exists \bx_1,\ldots,\bx_k \in \R^d, \forall \bx \in A, \exists i\in [ k ], \lno{ \bx-\bx_i}\le r
	} .
\]
if $A$ is nonempty, zero otherwise.
Then,
\begin{align*}
&
    \int_\cX \frac{\mathrm{d}\bx}{\brb{ f(\bxs) - f(\bx) + \e }^d}
\le
    \frac{\vol (\cX_\e) }{\e^d}
    + \sum_{k=1}^{\me}\frac{ \vol \brb{ \cX_{(\e_k ,\, \e_{k-1}]} } }{ (\e_k + \e)^d}
\\
&
\qquad\qquad
\le
    \frac{\cM \brb{ \cX_{\e} ,\, \frac{\e}{L} } \cdot \vol \brb{ \ball (1,\bzero) } \brb{ \frac{\e}{L} }^d }{ \e^d }
\\
&
\qquad\qquad\qquad
    + \sum_{k=1}^{\me}\frac{ \cM \brb{ \cX_{(\e_k ,\, \e_{k-1}]} ,\, \frac{\e_k }{L} } \cdot \vol \brb{ \ball (1,\bzero) } \brb{ \frac{\e_k }{L} }^d }{ \e_k ^d }
\\
&
\qquad\qquad
\le
    \frac{\vol \brb{ \ball (1,\bzero) }}{L^d} \lrb
    {
        \cN \lrb{ \cX_{\e}, \frac{\e}{L} } + 
        \sum_{k=1}^{\me} \cN \lrb{ \cX_{(\e_k ,\, \e_{k-1}]} ,\, \frac{\e_k }{L} }
    }
\\
&
\qquad\qquad
=
    \frac{\vol \brb{ \ball (1,\bzero) }}{L^d} 
    \SC(f,\e)
    \;,
\end{align*}
where the first inequality follows by lower bounding $f(\bxs) - f$ with its infimum on the partition, the second one by dropping $\e>0$ from the second denominator and upper bounding the volume of a set with the volume of the balls of a smallest $\nicefrac{\e_k}{L}$-covering, and the last one by the fact that covering numbers are always smaller than packing numbers (we recall this known result in \Cref{sec:lemmaPacking}, \cref{eq:wainwright}).
This proves the first part of the theorem.

For the second one, we have
\begin{align*}
&
    \int_\cX \frac{\mathrm{d}\bx}{\brb{ f(\bxs) - f(\bx) + \e }^d}
\ge
    \frac{\vol (\cX_{\e}) }{(\e + \e)^d}
    + \sum_{k=1}^{\me}\frac{ \vol \brb{ \cX_{(\e_k ,\, \e_{k-1}]} } }{ (\e_{k-1} + \e)^d}
\\
&
\qquad
\ge
    \frac{1}{2^d}
    \frac{\vol (\cX_{\e}) }{\e^d}
    + 
    \frac{1}{2^d}
    \sum_{k=1}^{\me}\frac{ \vol \brb{ \cX_{(\e_k ,\, \e_{k-1}]} } }{ \e_{k-1}^d}
\\
&
\qquad
\ge
    \frac 8 {15} \frac{1}{8^d} \lrb{
        \frac{ \vol \brb{ \cX_{2\e}} }{ \e^d }
        +
        \sum_{k=1}^m \frac{ \vol \brb{ \cX_{( \fracc{\e_k}{2}, \,  2 \e_{k-1} ]} } }{ \e_{k-1}^d }
    }
\\
&
\qquad
\ge
    \frac 8 {15}
    \frac
    { 
        \gamma \vol\lrb{ \frac{\e}{2L} \ball (1, \bzero) } \cN\lrb{\cX_{\e} ,\, \frac{\e}{L}}
    }
    {
        (8 \, \e)^d
    }
    +
\\
&
\qquad\qquad
    \frac 8 {15}
    \sum_{k=1}^{\me} \frac
    { 
        \gamma \vol\lrb{ \frac{\e_k}{2L} \ball (1, \bzero) } \cN\lrb{\cX_{(\e_k ,\, \e_{k-1}]} ,\, \frac{\e_k}{L}} 
    }
    { 
        (8 \, \e_{k-1} )^d
    }
\\
&
\qquad
\ge
    \frac 8 {15} \frac{\gamma \vol \brb{ \ball (1, \bzero) }}{(16L)^d}
    \, \cN\lrb{\cX_{\e} ,\, \frac{\e}{L}}
\\
&
\qquad\qquad
    + 
    \frac 8 {15} \frac{\gamma \vol \brb{ \ball (1, \bzero) }}{(32L)^d}
    \sum_{k=1}^{\me} 
    \cN\lrb{\cX_{(\e_k ,\, \e_{k-1}]} ,\, \frac{\e_k}{L}}
    \;,
\end{align*}
where the first inequality follows by upper bounding $f(\bxs) - f$ with its supremum on the partition, the second one by $\e \le \e_{k-1}$ (for all $k \in [\me+1]$), the third one by lower bounding the sum of disjoint layers with that of overlapping ones (proved in the appendix, \Cref{l:overlap}), the fourth one by the elementary inclusions 
$\cX_{2 \e } \supseteq \cX_{\frac{3}{2} \e}$
and
$\cX_{\left(\frac{1}{2} \e_k ,\, 2\e_{k-1}\right]} \supseteq \cX_{\left(\frac{1}{2} \e_k ,\, \frac{3}{2} \e_{k-1}\right]}$ (for all $k \in [\me]$) followed by a relationship between packing numbers and volumes (proved in the appendix, \Cref{p:pack-vol}) that holds under \Cref{ass:geomCx}, and the last one by $\e_{k-1}\le 2\e_k$ (for all $k \in [\me]$).
\end{proof}

\section{Missing proofs of Section \ref{s:upperbounds-alpha}}
\label{s:missing-upperbounds}

In this section, we include all missing details from \Cref{s:upperbounds-alpha}.

\subsection{Useful lemmas: general version}
\label{s:lemmas-alpha}

In this section, we extend the lemmas we proved in \Cref{s:lemmas} to the case of the \apiya{} algorithm run in a setting with perturbed observations.

\begin{lemma}
\label{l:fh-alpha}
Assume that $f\colon \cX \to \R$ is $L$-\lip{} around a maximizer $\bxs$ with respect to a norm $\lno \cdot$ (\Cref{ass:base}),
queries are corrupted by additive perturbations with absolute value bounded by some $\alpha\ge 0$,
and the \apiya{} (\Cref{a:acpiya}) is run with input $L$, $\lno \cdot$, $\bx_1$, $\eta$.
Then, for all $k \ge 1$, the proxy function 
$
    \fh_{k}(\cdot)
=
    \min_{i \in [k]} \bcb{ g(\bx_i)+L \lno{ \bx_i-(\cdot) } }
$
is $L$-Lipschitz and satisfies
\[
	\fh_k(\bxs) \ge f(\bxs) - \alpha
\qquad \text{ and } \qquad 
	\fh_k(\bx_j) \le f(\bx_j) + \alpha \;, \quad \forall j\in[k] \;.
\]
\end{lemma}	

\begin{proof}
Fix any $k\ge 1$. 
The fact that $\fh_k$ is $L$-\lip{} is straightforward. 
Moreover, by the $L$-\lipn{} of $f$ around $\bxs$, we get
\begin{align*}
    \fh_{k}(\bxs)
&
=
    \min_{i \in[k]} \Bcb{ f(\bx_i) + \alpha_i + L \lno{ \bx_i-\bxs } }
\\
&
\ge
    \min_{i \in[k]} \Bcb{ \brb{ f(\bxs) - L \lno{ \bx_i-\bxs } } - \alpha + L \lno{ \bx_i-\bxs } }
=
    f(\bxs) - \alpha \;.
\end{align*}
Furthermore, for all $j \in [k]$, the definition of $\fh_k(\bx_j)$ implies immediately
\[
	\fh_k(\bx_j) 
= 
	\min_{i \in[k]} \Bcb{ f(\bx_i) + \alpha_i + L \lno{ \bx_i-\bx_j } }
\le
	f(\bx_j) + \alpha + L \lno{ x_j - x_j }
=
	f(\bx_j) + \alpha \;. \qedhere
\]
\end{proof}
A first immediate consequence of \Cref{l:fh-alpha} is that the certified version of the \piya{} is a certified algorithm.

\begin{lemma}
Assume that $f\colon \cX \to \R$ is $L$-\lip{} around a maximizer $\bxs$ with respect to a norm $\lno \cdot$ (\Cref{ass:base}),
queries are corrupted by additive perturbations with absolute value bounded by some known $\alpha\ge 0$,
and the \AcpiyA{} (\Cref{a:acpiya}) is run with input $L$, $\lno \cdot$, $\bx_1$, $\eta$, $\alpha$.
Then, for all $n\ge 1$, $f(\bxs)-f(\bxs_n) \le \xi_n$.
\end{lemma}

\begin{proof}
By \Cref{l:fh-alpha} and the definitions of $f^\star_n \coloneqq g(\bxs_n)$, $\fhs_n \coloneqq \fh_n(\bx_{n+1}) \ge \sup \fh_n - \eta$, and $\xi_n \coloneqq \widehat f^\star_{n} - f^\star_{n} + (2\alpha+\eta)$, for all $n\ge 1$, we get
$
    f(\bxs) - f(\bxs_n) 
\le
    \brb{ \fh_n(\bxs) + \alpha} - \brb{f^\star_n -\alpha}
\le
    \xi_n
$.
\end{proof}

Another important implication of \Cref{l:fh-alpha} is that if the \piya{} observes $f$ at a $\Delta$-suboptimal point $\bx_i$, then all the next query points $\bx_j$ are at least $(\Delta/L)$-distant from $\bx_i$. 
In other words, the algorithm does not explore too much in suboptimal regions.

\begin{lemma}
\label{l:good-exploration-alpha}
Assume that $f\colon \cX \to \R$ is $L$-\lip{} around a maximizer $\bxs$ with respect to a norm $\lno \cdot$ (\Cref{ass:base}),
queries are corrupted by additive perturbations with absolute value bounded by some $\alpha\ge 0$,
and the \apiya{} (\Cref{a:acpiya}) is run with input $L$, $\lno \cdot$, $\bx_1$, $\eta$.
Fix any $\Delta>0$ and assume that there exists $i \ge 1$ such that the $i$-th query point $\bx_i$ belongs to $\cX_{\Delta}^c$. 
Then, for all $j > i$, the $j$-th query point satisfies
\[
	\lno{ \bx_j-\bx_i }
>
    \frac {\Delta - 2\alpha - \eta} L \;.
\]
\end{lemma} 

\begin{proof}
Fix any $j>i$. 
Then, using the fact that $\bx_j$ has been selected as the $j$-th query point (in the first inequality), \Cref{l:fh-alpha} (in the second and fourth), and $\bx_i \in \cX_\Delta^c$ (in the third), we get
\[
	\fh_{j-1}(\bx_j)
\ge 
	\fh_{j-1}(\bxs) - \eta
\ge 
	f(\bxs) - \alpha - \eta
>
	f(\bx_i) + \Delta - \alpha - \eta
\ge 
	\fh_{j-1}(\bx_i) + \Delta - 2\alpha - \eta \;.
\]
Since $\fh_{j-1}$ is $L$-\lip{} (by \Cref{l:fh-alpha}), we can thus conclude that
$
    L \lno{ \bx_j-\bx_i} 
\ge 
    \babs{ \fh_{j-1}(\bx_j)-\fh_{j-1}(\bx_i) }
>
    \Delta - 2\alpha - \eta
$.
\end{proof}

As a final corollary to \Cref{l:fh-alpha}, we show that the \AcpiyA{} automatically adapts its exploration to all possible approximation scales at the same time. 
In words, for any accuracy $\e$, until the algorithm can certify that the approximation error is $\e+2\alpha+\eta$ or better (i.e., until the error certificate is smaller than or equal to $\e+2\alpha+\eta$), it only queries points that are at least $(\e-2\alpha)/L$-distant from all past query points.
The strength of this statement comes from the fact that the algorithm \emph{does not} take an accuracy $\e$ as a parameter and the result is consequently not proved for a specific and known $\e$: it holds for all $\e$ simultaneously.

\begin{lemma}
\label{l:spread-exploration-alpha}
Assume that $f\colon \cX \to \R$ is $L$-\lip{} around a maximizer $\bxs$ with respect to a norm $\lno \cdot$ (\Cref{ass:base}),
queries are corrupted by additive perturbations with absolute value bounded by some known $\alpha\ge 0$,
and the \CpiyA{} (\Cref{a:acpiya}) is run with input $L$, $\lno \cdot$, $\bx_1$, $\alpha$. 
Fix any $\e > 0$ and let $\tau$ be the first time where the algorithm returns an error certificate $\xi_\tau \le \e + 2\alpha+\eta$.
Then, for all distinct $i,j \in [\tau]$, 
\[
    \lno{ \bx_i - \bx_j } 
> 
    \frac{\e - 2\alpha}{L}
    \;.
\]
\end{lemma}

\begin{proof}
Without loss of generality, assume $1 \le i < j \le \tau$. 
Then, by the definition of $\fh_{j-1}^\star$ (first equality), that of $\xi_{j-1}$ together with the assumption $\xi_{j-1} > \e + 2\alpha+\eta$ (first inequality), the definition of $f_{j-1}^\star$ (second), and \Cref{l:fh-alpha} (last), we get
\[
    \fh_{j-1}(\bx_j)
=
    \brb{  \fh_{j-1}^\star - f_{j-1}^\star }
    +
    f_{j-1}^\star
>
    \e + f_{j-1}^\star
\ge
    \e + f(\bx_i) - \alpha
\ge
    \e + \fh_{j-1}(\bx_i) - 2\alpha\;.
\]
Since $\fh_{j-1}$ is $L$-\lip{} (by \Cref{l:fh-alpha}), we can thus conclude that
$
    L \lno{ \bx_j-\bx_i} 
\ge 
    \babs{ \fh_{j-1}(\bx_j)-\fh_{j-1}(\bx_i) }
>
    \e - 2\alpha
$.
\end{proof}

The previous result shows that $\xi_n$ serves a synergistic double-role in the \AcpiyA{}.
When it is large, it guarantees that the queries are sufficiently sparse, and as soon as it becomes small, it certificates that the recommendation $\bx_n$ is sufficiently good.

\subsection{Proofs of Theorems \ref{t:SNC-upper-alpha} and \ref{t:SC-upper-alpha}}

\begin{proof}[Proof of \Cref{t:SNC-upper-alpha}.]
By definition of $\e_k^{\alpha,\eta}$, we can partition the set $\cX_{\e +(2\alpha+\eta)}^c = \cX_{(\e +(2\alpha+\eta), \, \eo]}$ of $(\e+2\alpha+\eta)$-suboptimal points into $\me$ layers
\[
    \cX_{\e +(2\alpha+\eta)}^c = \bigcup_{k = 1}^{\me} \cX_{(\e_k^{\alpha,\eta} ,\, \e_{k-1}^{\alpha,\eta}]} \;.
\]
Therefore, for any $i \ge 1$, if $\bx_i\in \cX_{\e + (2\alpha+\eta)}^c$, then there exists a unique $k\in [\me]$ such that $\bx_i \in \cX_{(\e_k^{\alpha,\eta} ,\, \e_{k-1}^{\alpha,\eta}]}$.
Since the \piya{} does not explore too much in suboptimal regions, \Cref{l:good-exploration-alpha} and the definition of packing numbers imply that for all $k\in [\me]$, the maximum number of times that a query point $\bx_i$ is chosen in $\cX_{(\e_k^{\alpha,\eta} ,\, \e_{k-1}^{\alpha,\eta}]} \s \cX_{\e_k^{\alpha,\eta}}^c$ is upper bounded by 
$
    \cN \lrb{ \cX_{(\e_k^{\alpha,\eta} ,\, \e_{k-1}^{\alpha,\eta}]} ,\, \frac{\e_k^{\alpha,\eta} - 2\alpha-\eta}{L} }
$.
Hence
\begin{multline*}
    \Babs{ \bcb{ i\in\{1,2,\dots\} : \bx_i\in\cX_{\e+(2\alpha+\eta)}^c } }
=
    \sum_{k=1}^{\me} \Babs{ \bcb{ i\in\{1,2,\dots\} : \bx_i\in \cX_{(\e_k^{\alpha,\eta} ,\, \e_{k-1}^{\alpha,\eta}]} } }
\\
\le
    \sum_{k=1}^{\me}\cN \lrb{ \cX_{(\e_k^{\alpha,\eta} ,\, \e_{k-1}^{\alpha,\eta}]} ,\, \frac{\e_k^{\alpha,\eta}-2\alpha-\eta}{L} }
=
    \SNC(f,\e) \;.    
\end{multline*}
Therefore, if
$
    n
\ge
    1 + \SNC^{\alpha,\eta}(f,\e)
$,
then there exists $i\ge 1$ such that $\bx_i\in\cX_{\e+(2\alpha+\eta)}$, which implies (by definition of $\bxs_n$) that 
$
    f(\bxs_n)
\ge 
    f(\bx_i) - 2\alpha
\ge 
    f(\bxs) - \e - (2\alpha+ \eta) - 2\alpha
$ 
and concludes the proof.
\end{proof}

\begin{proof}[Proof of \Cref{t:SC-upper-alpha}.]
By definition of $\e_k^{\alpha,\eta}$, we can partition $\cX$ into the $\me+1$ sets
\[
    \cX 
= 
    \cX_{\e_{\me}^{\alpha,\eta}}
    \cup
    \bigcup_{k = 1}^{\me} \cX_{(\e_k^{\alpha,\eta} ,\, \e_{k-1}^{\alpha,\eta}]} \;.
\]
Let $\tau$ be the first time where the algorithm returns an error certificate $\xi_\tau \le \e  + 2\alpha+\eta$.
Then, for any $i \in [\tau]$, either $\bx_i \in \cX_{\e_{\me}^{\alpha,\eta}}$ or there exists a unique $k\in [\me]$ such that $\bx_i \in \cX_{(\e_k^{\alpha,\eta} ,\, \e_{k-1}^{\alpha,\eta}]}$.
Since the \CpiyA{} automatically adapts its exploration to all possible approximation scales at the same time, \Cref{l:spread-exploration-alpha} and the definition of packing numbers imply that the maximum number of times that a query point $\bx_i$ is chosen in $\cX_{\e_{\me}^{\alpha,\eta}}$ (up to and including time $\tau$) is upper bounded by 
$
    \cN \lrb{ \cX_{\e_{\me}^{\alpha,\eta}} ,\, \frac{\e_{\me}^{\alpha,\eta}-2\alpha}{L} }
$.
Similarly, since the \CpiyA{} does not explore too much in suboptimal regions, \Cref{l:good-exploration-alpha} and the definition of packing numbers imply that for all $k\in [\me]$, the maximum number of times that a query point $\bx_i$ is chosen in $\cX_{(\e_k^{\alpha,\eta} ,\, \e_{k-1}^{\alpha,\eta}]} \s \cX_{\e_k^{\alpha,\eta}}^c$ is upper bounded by 
$
    \cN \lrb{ \cX_{(\e_k^{\alpha,\eta} ,\, \e_{k-1}^{\alpha,\eta}]} ,\, \frac{\e_k^{\alpha,\eta}-2\alpha-\eta}{L} }
$.
Hence
\begin{multline*}
    \tau
=
    \Babs{ \bcb{ i\in[\tau] : \bx_i\in\cX_{\e_{\me}^{\alpha,\eta}} } }
    +
    \sum_{k=1}^{\me} \Babs{ \bcb{ i\in[\tau] : \bx_i\in \cX_{(\e_k^{\alpha,\eta} ,\, \e_{k-1}^{\alpha,\eta}]} } }
\\
\le
    \cN \lrb{ \cX_{\e_{\me}^{\alpha,\eta}} ,\, \frac{\e_{\me}^{\alpha,\eta}-2\alpha}{L} }
    +
    \sum_{k=1}^{\me}\cN \lrb{ \cX_{(\e_k^{\alpha,\eta} ,\, \e_{k-1}^{\alpha,\eta}]}, \frac{\e_k^{\alpha,\eta}-2\alpha-\eta}{L} }
=
    \SC^{\alpha,\eta}(f,\e)\;.    
\end{multline*}
Therefore, noting that $k\mapsto \xi_k$ is non-increasing (because $\xi_k +2\alpha+\eta= \fh_k^\star - f_k^\star + 2\alpha+\eta$, with $k\mapsto \fh_k^\star$ non-increasing and $k\mapsto f_k^\star$ non-decreasing), if
$
    n
\ge
    \SC(f,\e)
$,
then
$
    n
\ge
    \tau
$,
which in turn yields
$\xi_n \le \xi_\tau \le \e + 2\alpha+\eta$.
\end{proof}

\end{document}